%% file: main.tex
\documentclass[10pt,twocolumn,letterpaper]{article}

\usepackage[]{cvpr} 
\usepackage{arydshln}
\usepackage{times}
\usepackage{epsfig}
\usepackage{graphicx}
\usepackage{amsmath}
\usepackage{amssymb}
\usepackage{enumitem}
\usepackage{booktabs}

\input{preamble}

\usepackage{rotating}
\makeatletter
\DeclareRobustCommand\onedot{\futurelet\@let@token\@onedot}
\def\@onedot{\ifx\@let@token.\else.\null\fi\xspace}

\usepackage{multirow}
\usepackage{algorithm}
\usepackage[noend]{algorithmic}
\usepackage{graphicx, amsmath, amssymb, caption, subcaption, multirow, overpic, textpos}

\def\eg{\emph{e.g}\onedot} 
\def\ie{\emph{i.e}\onedot} 
 
 \def\vs{\emph{vs}\onedot}
\def\wrt{w.r.t\onedot}

\newcommand{\figref}[1]{Fig\onedot~\ref{#1}}

\newcommand{\secref}[1]{Sec\onedot~\ref{#1}}
\newcommand{\tabref}[1]{Tab\onedot~\ref{#1}}

\usepackage{pifont}

\usepackage{xcolor}

\definecolor{siglipcolor}{gray}{.75}
\definecolor{gray}{gray}{.75}
\definecolor{baselinecolor}{gray}{.9}
\newcommand{\baseline}[1]{\cellcolor{baselinecolor}{#1}}

\definecolor{settingcolor}{gray}{0.9}

\definecolor{lightblue}{rgb}{0.85,0.9,1}
\newcommand{\movittable}[1]{\cellcolor{lightblue}{#1}}

\newlength\savewidth\newcommand\shline{\noalign{\global\savewidth\arrayrulewidth
  \global\arrayrulewidth 1pt}\hline\noalign{\global\arrayrulewidth\savewidth}}
\newcommand{\tablestyle}[2]{\setlength{\tabcolsep}{#1}\renewcommand{\arraystretch}{#2}\centering\footnotesize}

\definecolor{cvprblue}{rgb}{0.21,0.49,0.74}
\usepackage[pagebackref,breaklinks,colorlinks,citecolor=cvprblue]{hyperref}

\def\paperID{150} 
\def\confName{CVPR}
\def\confYear{2024}

\newcommand{\modelname}{ViTamin\xspace}

\newcommand{\ovdetmodelname}{Sliding F-ViT\xspace}

\newcommand{\ovsegmodelname}{Sliding FC-CLIP\xspace}

\title{ViTamin: Designing Scalable Vision Models in the Vision-Language Era}

\author{
	Jieneng Chen$^{1}$\footnotemark[1]
	\;\; Qihang Yu$^{2}$\footnotemark[1]
	\;\; Xiaohui Shen$^2$
	\;\; Alan Yuille$^1$
	\;\; Liang-Chieh Chen$^2$ \\
	$^1$Johns Hopkins University \;\; $^2$ByteDance  \;\; *equal contribution
 \\
\url{https://beckschen.github.io/vitamin.html}  
}

\begin{document}
\maketitle

\input{sec/0_abstract} 
\input{sec/1_intro}
\input{sec/2_related}
\input{sec/3_method}

\input{sec/5_experiment}   
\input{sec/6_conclusion}

\newpage

\input{sec/X_suppl}

{
    \small
    \bibliographystyle{ieeenat_fullname}
    \bibliography{main}
}


\end{document}

%% file: preamble.tex
\usepackage[table,dvipsnames]{xcolor}


%% file: sec/0_abstract.tex
\begin{abstract}

Recent breakthroughs in vision-language models (VLMs) start a new page in the vision community.
The VLMs provide stronger and more generalizable feature embeddings compared to those from ImageNet-pretrained models, thanks to the training on the large-scale Internet image-text pairs.
However, despite the amazing achievement from the VLMs, vanilla Vision Transformers (ViTs) remain the default choice for the image encoder.
Although pure transformer proves its effectiveness in the text encoding area, it remains questionable whether it is also the case for image encoding, especially considering that various types of networks are proposed on the ImageNet benchmark, which, unfortunately, are rarely studied in VLMs. Due to small data/model scale, the original conclusions of model design on ImageNet can be limited and biased. In this paper, we aim at building an evaluation protocol of vision models in the vision-language era under the contrastive language-image pretraining (CLIP) framework. We provide a comprehensive way to benchmark different vision models, covering their zero-shot performance and scalability in both model and training data sizes. To this end, we introduce \modelname, a new vision models tailored for VLMs.
\modelname-L significantly outperforms ViT-L by 2.0\% ImageNet zero-shot accuracy, when using the same publicly available DataComp-1B dataset and the same OpenCLIP training scheme.
\modelname-L presents promising results on 60 diverse benchmarks, including classification, retrieval, open-vocabulary detection and segmentation, and large multi-modal models.
When further scaling up the model size, our ViTamin-XL with only 436M parameters attains 82.9\% ImageNet zero-shot accuracy, surpassing 82.0\% achieved by EVA-E that has ten times more parameters (4.4B).
\end{abstract}

%% file: sec/1_intro.tex
\section{Introduction}
\label{sec:intro}

\begin{figure}
  \centering
  \includegraphics[width=0.91\linewidth]{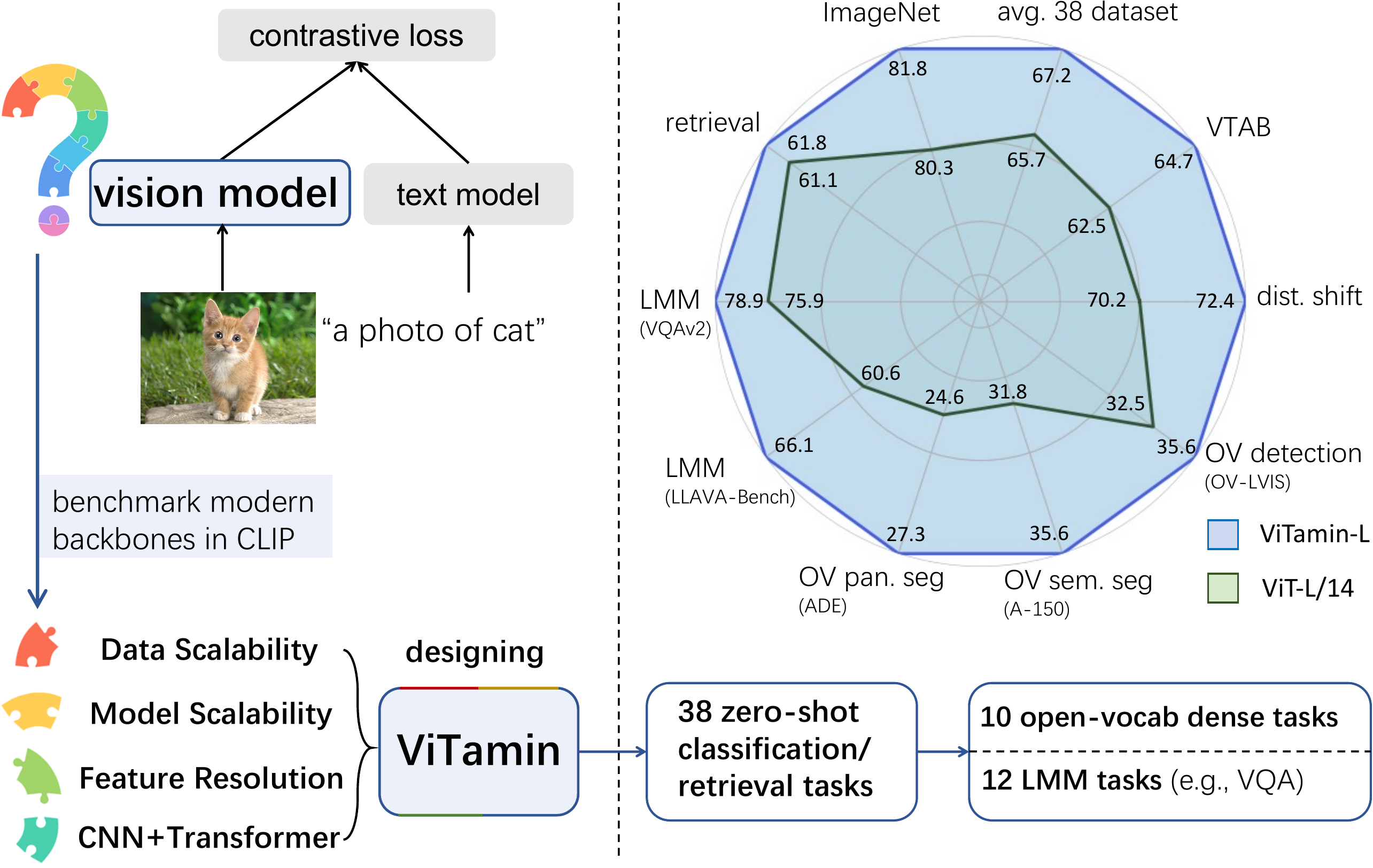}
  \caption{\textbf{Practices of designing scalable vision models in the vision-language era.} We benchmark modern vision models with various model and data scales under CLIP setting using DataComp-1B~\cite{gadre2023datacomp}, leading to findings about data and model scalability, feature resolution, and hybrid architecture, which motivate us to develop \modelname for VLM. \modelname-L achieves superior zero-shot performance over ViT-L/14~\cite{li2023clipa} on ImageNet~\cite{russakovsky2015imagenet} and average 38 datasets~\cite{gadre2023datacomp}, and advances a suite of 22 downstream tasks for Open-Vocabulary (OV) detection~\cite{wu2023clipself} and segmentation~\cite{yu2023convolutions}, and Large Multi-modal Model (LMM) tasks~\cite{liu2023improvedllava}.
  }
  \label{fig:teaser}
  \vspace{-5mm}
\end{figure}

The past decades have witnessed significant progress in computer vision, like visual recognition tasks.
The advent of AlexNet~\cite{krizhevsky2012imagenet} marked a significant milestone, catalyzing the extensive evolution and dominance of Convolutional Neural Networks (ConvNets)~\cite{lecun1998gradient,he2016deep,huang2017densely,long2015fully,chen015deeplabv1, girshick2015fast,he2017mask,liu2022convnet} in computer vision.
More recently, with the debut of Vision Transformer~\cite{vaswani2017attention, dosovitskiy2020image}, a growing number of transformer-based architectures~\cite{wang2021pyramid,yu2021glance,liu2021swin,dai2021coatnet,yang2023moat,tu2022maxvit} have shown great potential to surpass the prior ConvNet counterparts.

The rapid advancement of neural network design in computer vision can be attributed to a combination of factors.
Among them, an important factor is the well-established benchmarks, allowing the community to  examine the developments in a standardized way.
Particularly, ImageNet~\cite{russakovsky2015imagenet} has become the de facto testing ground for new vision models.
It not only sets a standard benchmark for visual recognition, but also serves as a mature pre-training dataset for transferring the network backbone to a variety of downstream tasks (\eg, detection and segmentation)~\cite{lin2014microsoft,long2015fully,chen2017deeplab, he2017mask,chen2018encoder,kirillov2019panoptic,cheng2020panoptic,wang2021max,yu2022cmt,yu2022k,sun2024remax,yang2024polymax}.

Recently, the emergence of vision-language models (VLMs)~\cite{radford2021learning, jia2021scaling} has changed the paradigm by leveraging the pre-training schedule on the extremely large scale noisy Internet data up to billions of image-text pairs~\cite{schuhmann2022laion}, much larger than the ImageNet scale.
VLMs not only produce strong and generalizable features~\cite{radford2021learning, jia2021scaling}, but also excel in zero-shot  downstream tasks~\cite{rombach2022high,minderer2022simple,ghiasi2022scaling,kuo2022f,yu2023convolutions,liu2023visual,zhu2023minigpt}.
However, unlike the ImageNet benchmark, where many types of neural networks are designed and blossomed~\cite{krizhevsky2012imagenet,simonyan2014very,he2016deep,huang2017densely,howard2017mobilenets,tan2019efficientnet}, the existing VLMs mostly employ the vanilla Vision Transformer (ViT) architecture~\cite{dosovitskiy2020image} \footnote{with only a few exceptions, \eg, ConvNeXt~\cite{liu2022convnet} by OpenCLIP~\cite{ilharco_gabriel_2021_5143773}.}, and the recent benchmark DataComp~\cite{gadre2023datacomp} focuses on the data curation under the common (yet unverified) belief that ViTs scale much better than any other architectures in this vision-language era~\cite{dehghani2023scaling,li2023inverse} and thus ViT is all we need.

The current trend can be characterized by several key observations: 
(1)  The high computational demand, requiring extensive resources~\cite{ilharco_gabriel_2021_5143773} for months, is a significant barrier for advancing VLMs~\cite{radford2021learning}, limiting exploring diverse vision models.
(2)  Traditional vision models are mainly optimized for the ImageNet benchmark, which may not scale well for larger datasets~\cite{schuhmann2022laion, gadre2023datacomp}, unlike purely transformer-based architectures~\cite{vaswani2017attention}  that have proven scalable in language tasks~\cite{openai2023gpt,touvron2023llama2} and are now being adopted for VLMs as image encoders~\cite{dosovitskiy2020image}.
(3)  Current VLM benchmarks focus on zero-shot classification/retrieval tasks~\cite{gadre2023datacomp}, with a notable lack of downstream tasks involving open-vocabulary dense prediction~\cite{zareian2021open,ghiasi2022scaling,xu2021simple,zhou2022extract,kuo2022f,ding2023open,xu2023open,yu2023convolutions,deng2024coconut}, as well as a gap in assessing Large Multi-modal Models (LMMs)~\cite{li2022blip,liu2023visual, liu2023improvedllava,zhu2023minigpt}.

In this paper, we aim to address the aforementioned issues with practices as shown in \figref{fig:teaser}.
To begin with, we establish a new test bed for designing vision models under the CLIP framework~\cite{radford2021learning,jia2021scaling} using the DataComp-1B dataset~\cite{gadre2023datacomp}, which is one of the largest publicly available datasets with high quality. Specifically, we employ two training protocols: \textit{short schedule} for fast benchmarking vision models across model and data scales, and \textit{long schedule} for training best performing vision models. With the \textit{short schedule}, we re-benchmark state-of-the-art vision models found on ImageNet settings for VLMs. Particularly, we select ViT~\cite{dosovitskiy2020image}, ConvNeXt~\cite{liu2022convnet}, CoAtNet~\cite{dai2021coatnet}, as representatives for pure Transformer, pure ConvNet, and hybrid architecture, respectively. We combine various model scales and data scales to provide a comprehensive evaluation towards different architectures, revealing several  critical findings.
First, increasing data scales improves all vision models across all model sizes, while ViT scale slightly better than others in terms of model parameters. Second, the final resolution of the extracted features affects prediction performance. Third, CoAtNet performs better than ViT and ConvNeXt in general, though it is hard to scale up CoAtNet-4 to billions of data due to computational constraints. 

Those findings motivate us to develop a new vision model, named \modelname tailored for VLM. \modelname is a 3-stage hybrid architectures, combining two stages of MBConv blocks with a final stage of Transformer blocks. This hybrid design leverages its Transformer stage to enhance data and model scalability, along with output stride of 16 to enjoy high feature resolution. As a result, \modelname-L outshines its ViT-L/14 counterpart~\cite{gadre2023datacomp} by +2.0\% zero-shot imageNet accuracy in identical OpenCLIP training scheme and identical 256 token length. When increasing feature resolution to 576 patches, \modelname-L further attains 81.8\% zero-shot imageNet accuracy, surpassing the prior art ViT-L/14 CLIPA-v2~\cite{li2023clipa} by +1.5\%. In average performance across 38 datasets, it not only exceeds ViT-L/14 counterpart~\cite{li2023clipa} by +1.5\%, but also outperforms the larger ViT-H/14 model~\cite{li2023clipa} by +0.4\% while having only half parameters. When further scaling up the model size, our ViTamin-XL with only 436M parameters attains 82.9\% ImageNet zero-shot accuracy, surpassing 82.0\% achieved by EVA-E (\ie, EVA-02-CLIP-E/14~\cite{sun2023eva}) that has ten times more parameters (4.4B). Furthermore, we introduce an effective training scheme Locked-Text Tuning (LTT), which guides the training of vision backbone with a frozen pretrained text encoder.  It enhances the small variant by +4.0\% and the base variant by +4.9\%  without any extra cost.

Our another intriguing observation is the prevailing emphasis on data filtering over vision architecture design in VLM. For instance, while the best DataComp challenge solution~\cite{yu2023devil} achieved only a +2.3\% gain, our \modelname  with LTT largely improves performance by +23.3\% on the same dataset size, without intensive data filtering. 
Finally, we introduce a suite of downstream tasks, including open-vocabulary detection and segmentation, and LMMs, for evaluating VLM-specific vision models. \modelname outperforms the ViT-L model, enhancing detector by +3.1\% on OV-LVIS and segmentor by +2.6\% on average 8 datasets, and excelling across 12 LMM benchmarks.
Notably, \modelname sets a new state-of-the-art on 7 benchmarks for open-vocabulary segmentation.

We aim for our findings to encourage a reevaluation of the current limitations in VLM designs and hope that our extensive benchmarking and evaluations will drive the development of more advanced vision models for VLMs.

%% file: sec/2_related.tex
\section{Related Work}

\textbf{Vision Backbone:}
On the ImageNet benchmark~\cite{russakovsky2015imagenet}, ConvNets~\cite{krizhevsky2012imagenet, simonyan2014very, szegedy2015going, he2016deep, huang2017densely, xie2017aggregated, sandler2018mobilenetv2, tan2019efficientnet, tan2021efficientnetv2, liu2022convnet, yu2023exploring} have been the dominant networks choice since the advent of AlexNet~\cite{krizhevsky2012imagenet}. Recently, the vision community has witnessed impressive emergence of the Transformer architecture~\cite{vaswani2017attention}, a trend that began with the widespread adoption of the ViT~\cite{dosovitskiy2020image} and its subsequent developments~\cite{ramachandran2019stand,wang2020axial,touvron2021training,fan2021multiscale,liu2021swin,wang2021pyramid,zhou2021deepvit,li2022mvitv2,yu2022metaformer,li2023uniformer}. Among these, hybrid architectures~\cite{graham2021levit,srinivas2021bottleneck,d2021convit,heo2021rethinking,wu2021cvt,dai2021coatnet,xiao2021early,mehta2021mobilevit,chen2022mobile,tu2022maxvit,li2022efficientformer,yang2023moat} combine Transformer self-attention with convolution, where CoAtNet~\cite{dai2021coatnet} particularly obtains impressive results on ImageNet.
Notably, MaX-DeepLab~\cite{wang2021max}, emerged as early as 2020, successfully developed a hybrid network backbone for dense pixel predictions, where the first two stages utilize residual bottleneck blocks~\cite{he2016deep}, followed by two subsequent stages employing axial attention~\cite{wang2020axial}. 
More recently, by leveraging the design practices of a Vision Transformer, a ResNet~\cite{he2016deep} can be modernized to ConvNeXt~\cite{liu2022convnet}, competing favorably with ViT.
Along the same direction, but not limited to the ImageNet scale, our work aims to develop a novel vision model for training with billions of data~\cite{gadre2023datacomp} in the vision-language era.

\textbf{Language-Image Pre-training:}
Language-image pre-training has seen significant advancements~\cite{jia2021scaling,radford2021learning,li2022blip,alayrac2022flamingo,yu2022coca,liu2023visual,Betker2023} with the emergence of LLMs~\cite{brown2020language, openai2023gpt,touvron2023llama}.
The huge progress can be attributed to the pre-training on an immense scale of noisy web-collected image-text data~\cite{schuhmann2022laion,gadre2023datacomp}, much larger than the ImageNet. Notably, CLIP~\cite{radford2021learning,jia2021scaling} generates strong image features and excels in zero-shot transfer learning~\cite{rombach2022high,minderer2022simple,ghiasi2022scaling,kuo2022f,yu2023convolutions,liu2023visual,zhu2023minigpt}, which make it an essential role in large multi-modal model~\cite{li2022blip,liu2023visual,chen2023pali,liu2023improvedllava}. CLIP has been improved by advanced training strategies including self-supervised learning~\cite{mu2022slip,li2023scaling}, efficient tuning~\cite{zhai2022lit,liu2023learning} and training~\cite{li2023inverse,li2023reclip, sun2023eva, wu2023tinyclip,zhai2023sigmoid}. 
These studies predominantly employ ViT~\cite{dosovitskiy2020image} as the only vision model.
As a result, the architectural design for the CLIP vision model has not been thoroughly investigated. Thus, we attempt to bridge the gap by developing a novel vision model for VLMs.

%% file: sec/3_method.tex
\section{Method}

In the section, we revisit the problem definition of CLIP and propose two training protocols (\textit{short} and \textit{long} schedules) on DataComp-1B (\secref{sec:clip_and_training}).
With \textit{short schedule}, we re-benchmark modern vision models found on ImageNet under the CLIP setting (\secref{sec:rebenchmark}).
We then introduce the proposed \modelname architecture design, motivated by the discoveries in the re-benchmarking results (\secref{sec:proposed_model}).

\subsection{CLIP and Training Protocols}
\label{sec:clip_and_training}

\textbf{CLIP Framework:} 
Given a batch of $N$ image-text pairs $\{\left(I_1, T_1\right), ..., \left(I_N, T_N\right)\}$ (where $I_i$ and $T_i$ denote image and text for $i_{th}$ pair), the objective of CLIP~\cite{radford2021learning}
learns to align the image embeddings $\mathbf{x}_i$ and text embeddings $\mathbf{y}_i$ for each pair.
Formally, the loss function is defined as follows:
\vspace{-2mm}
\begin{equation}
    -\frac{1}{2N} \sum_{i=1}^{N}(\underbrace{\log \frac{e^{ \mathbf{x}_i^{T}  \mathbf{y}_i / \tau} }{\sum_{j=1}^{N} e^{ \mathbf{x}_i^{T}  \mathbf{y}_j / \tau}}}_{\text {image to text  }}+\underbrace{\log \frac{e^{ \mathbf{y}_i^{T} \mathbf{x}_i / \tau}}{\sum_{j=1}^{N} e^{  \mathbf{y}_i^{T} \mathbf{x}_j / \tau}}}_{\text {text to image  }}),
    \label{equ:clip} \vspace{-2mm}
\end{equation}
where $\mathbf{x}_i=\frac{f\left(I_i\right)}{\left\|f\left(I_i\right)\right\|_2}$,  $\mathbf{y}_i=\frac{g\left(T_i\right)}{\left\|g\left(T_i\right)\right\|_2}$, and $\tau$ is a  temperature variable.
A vision model $f\left(.\right)$ and a text model $g\left(.\right)$ are trained to minimize the loss function. We focus on vision model design and use the text models from OpenCLIP~\cite{ilharco_gabriel_2021_5143773}.

\textbf{Training Protocols:}
We employ two training protocols: \textit{short schedule} and \textit{long schedule}.
The \textit{short schedule} is designed for efficiently benchmarking vision models up to 1 training epoch on DataComp-1B~\cite{gadre2023datacomp} (\ie, 1.28B seen samples).
As detailed in \tabref{tab:training_protocol}, given a descent amount of resources (\eg, 32 A100 GPUs), it takes less than two days to train a small ($\sim$25M parameters) model variant.
The \textit{long schedule} is designed for training the best performing models with up to 40B seen samples.

\subsection{Benchmarking Vision Models in CLIP Setting}
\label{sec:rebenchmark}
\begin{figure*}
  \vspace{-4mm}
  \centering
  \includegraphics[width=0.99\linewidth]{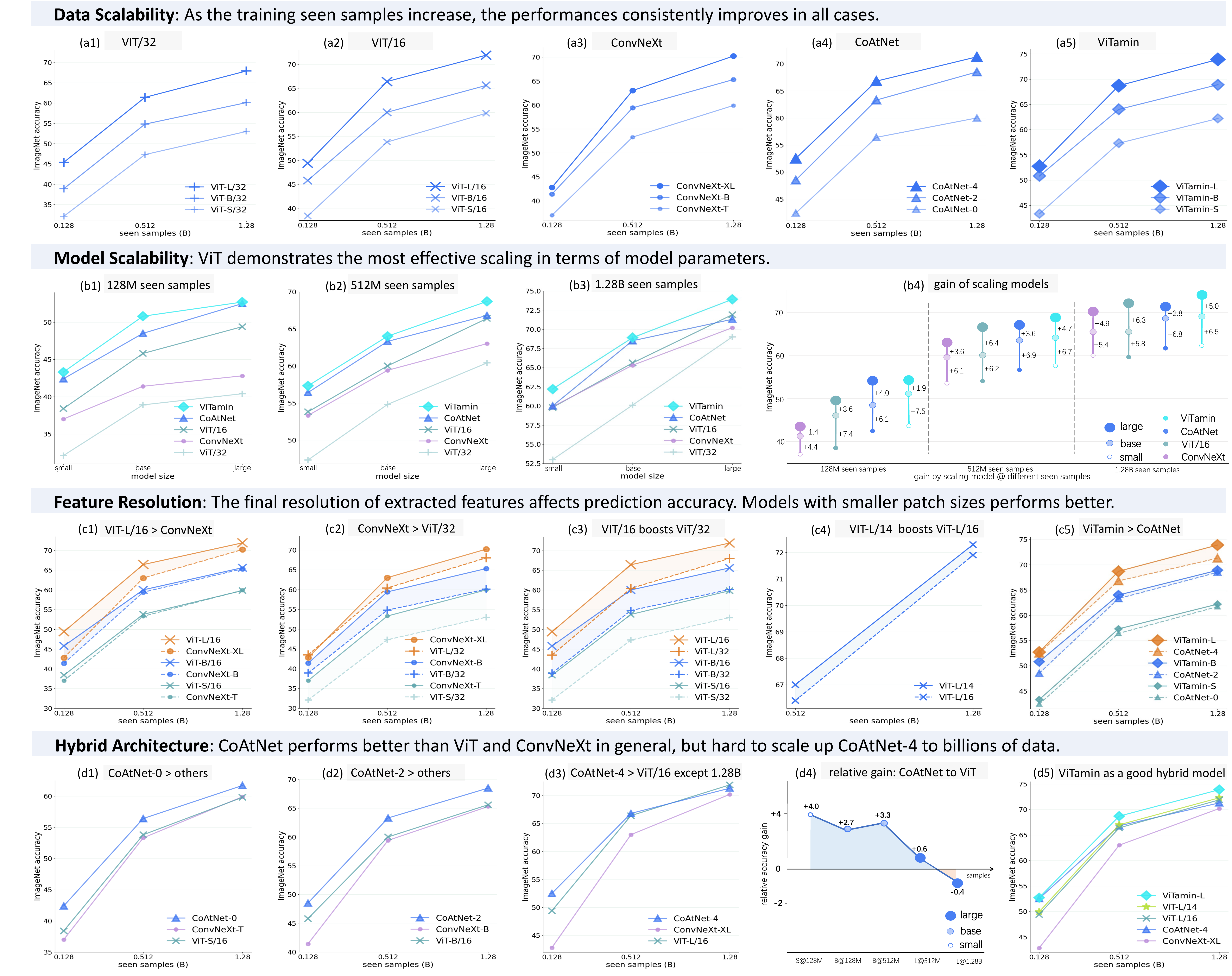}
  \caption{\textbf{Benchmarking vision models under CLIP setting on DataComp-1B}, including ViT (a pure Transformer), ConvNeXt (a pure ConvNet), and CoAtNet (a hybrid model).  We examine their scalability in terms of both data sizes (1st row) and model scales (2nd row), and further analyze the results from the aspects of feature resolution (3rd row) and hybrid architecture (4th row). 
  } 
  \label{fig:benchmark}
  \vspace{-4mm}
\end{figure*}

The \textit{short schedule} allows us to efficiently re-benchmark state-of-the-art vision models found on ImageNet under the CLIP setting using DataComp-1B.
The experimented models are ViT~\cite{dosovitskiy2020image} (a pure Transformer), ConvNeXt~\cite{liu2022convnet} (a pure ConvNet), and CoAtNet~\cite{dai2021coatnet} (a hybrid model).
We examine their scalability in terms of both model scales and data sizes.
Each vision model has sizes varying from small ($\sim$25M parameters), base ($\sim$85M) to large ($\sim$300M), while the data sizes range from 128M, 512M to 1.28B training seen samples (1 epoch is equal to 1.28B seen samples). 
The metric is zero-shot accuracy on ImageNet, supplemented by the results on the 38 tasks following DataComp~\cite{gadre2023datacomp}.
As shown in \figref{fig:benchmark}, we analyze the benchmarked results from four aspects, including data scalability, model scalability, feature resolution, and hybrid architecture. For simplicity, we use ``X@Y'' to denote the vision model X trained with Y seen samples. See appendix for numerical results.

\textbf{Data Scalability:}
When training seen samples increase from 128M to 1.28B, we observe a consistent trend of improvements across all model sizes and all vision models (\hyperref[fig:benchmark]{a1-a5}). Interestingly, ViT-S/16@512M (22M parameter) attains the zero-shot performance of 53.8\% on ImageNet, which is better than 45.8\% by ViT-B/16@128M (86M parameter). It shows the effectiveness of training large scale data that quadrupling training seen samples can be more impactful than quadrupling the number of model parameters.
Additionally, ViT-B/16@512M \& @1.28B significantly boost ViT-B/16@128M from 45.8\% to 60.0\% (+14.2\%) and 65.6\% (+19.8).

$\rightarrow$ \textit{As the training seen samples increase, the performances consistently improves in all cases.}

\textbf{Model Scalability:}
When the model sizes increase, the performances of all vision models are also boosted (\hyperref[fig:benchmark]{b1-b3}). However, we observe a different gain among them (\hyperref[fig:benchmark]{b4}). For example, ConvNeXt-XL@128M brings only +1.4\% gain over ConvNeXt-B@128M, while ViT-L/16@128M enhances ViT-B/16@128M by +3.6\%.  Given plenty of data, ViT still shows a better model scalability, especially scaling from base to large (\eg, +6.4\% for ViT \vs +3.6\% for both CoAtNet and ConvNeXt at 512M samples; +6.3\% for ViT \vs +2.8\% for CoAtNet and + 4.9\% for ConvNeXt at 1.28B samples). As a result, ViT shows the best scalability.

$\rightarrow$ \textit{ViT demonstrates the most effective scaling in terms of model parameters.}

\textbf{Feature Resolution:}
Across all model scales and data sizes, ConvNeXt performs better than ViT/32 but loses its advantage to ViT/16 (\hyperref[fig:benchmark]{c1 \& c2}).
This trend deviates significantly from what is observed in ImageNet era, where ConvNeXt consistently outperforms ViT/16 (also see our ImageNet-scale VLM experiments in appendix). We hypothesize that, comparing to ImageNet's object class label, the text in CLIP captures broader area of information, and thus is beneficial from higher feature resolution. Besides, ViT also benefits from using smaller patch sizes (thus high feature resolution) over larger path sizes (\hyperref[fig:benchmark]{c3 \& c4}).

$\rightarrow$ \textit{The final resolution of extracted features affects the prediction performance. ViT with smaller patch sizes outperforms ViT with larger patch sizes and ConvNeXt.}
\vspace{-0.5mm}

\textbf{Hybrid Architecture:}
We observe that ConvNeXt consistently lags behind ViT-\{S,B\}/16 and particularly ViT-L/14, suggesting that a pure ConvNet has limited capacity under the CLIP setting when presented with abundant of data (\hyperref[fig:benchmark]{d1-d3}).
By contrast, CoAtNet significantly surpasses both ViT and ConvNeXt (\eg, CoAtNet-2@1.28B has a remarkable +2.9\% and +3.2\% gain over ViT-B/16@1.28B and ConvNeXt-B@1.28B, respectively), indicating the effectiveness of hybrid models.
However, CoAtNet requires the most GPU memory; we can only train CoAtNet-4 with batch size 8k on 64 A100 GPUs, while all the other large models are trained with batch size 16k on 32 A100 GPUs. This affects CoAtNet's scalability in large variant.

$\rightarrow$ \textit{CoAtNet surpasses ViT and ConvNeXt in general, yet it is hard to scale up CoAtNet-4 to billions of data.}

\subsection{Novel Vision Transformer for Vision-Language} 
\label{sec:proposed_model}

Herein, we distill from the aforementioned observations, culminating in the proposed vision model, \modelname (\textbf{Vi}sion \textbf{T}r\textbf{A}nsfor\textbf{M}er for v\textbf{I}sion-la\textbf{N}guage), which notably takes the lead in the benchmarking results across all settings in \figref{fig:benchmark}.
To introduce \modelname, we commence by its macro-level network design (\secref{sec:macro}), followed by the micro-level block design (\secref{sec:micro}).
Finally, we develop a vision model family with a simple scaling rule (\secref{sec:scale}).

\begin{figure*}
  \vspace{-3mm}
  \centering
  \includegraphics[width=0.95\linewidth]{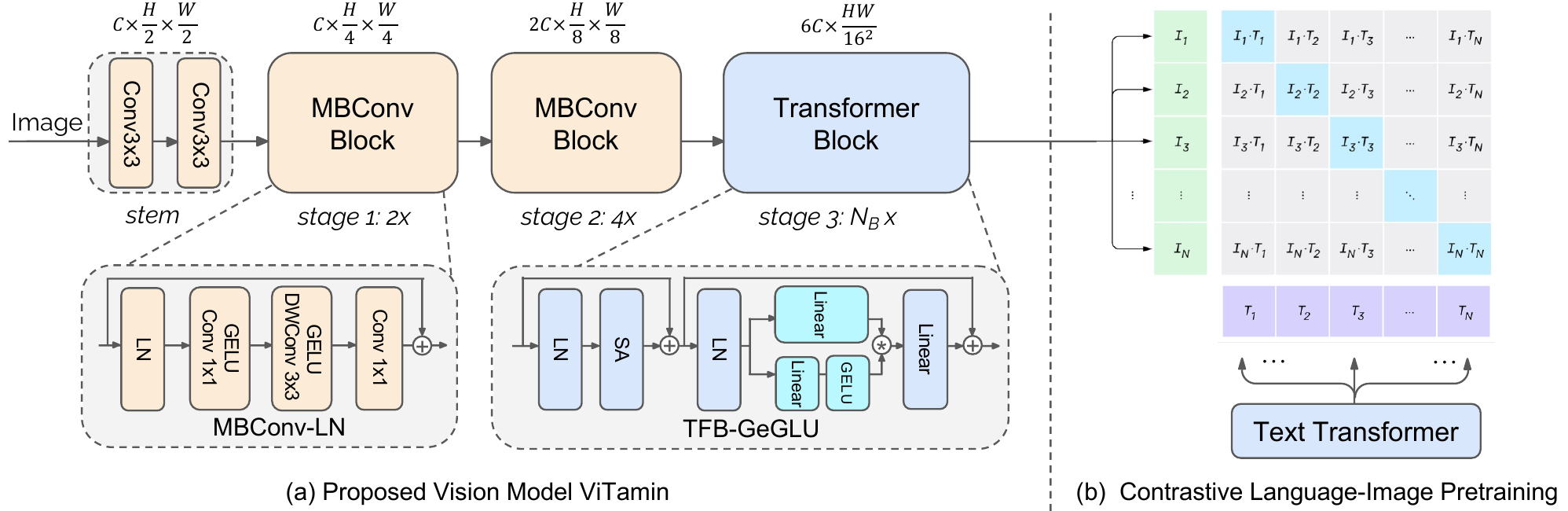}
  \caption{\textbf{Overview of \modelname architecture}. (a) \modelname begins with a convolutional stem, followed by Mobile Convolution Blocks (MBConv) in stage 1 and 2, and Transformer Blocks (TFB) in stage 3. The 2D input to the stage 3 is flattened to 1D. 
  For the \textit{macro-level} designs, the three-stage layout generates the final feature map with output stride 16, similar to ViT/16~\cite{dosovitskiy2020image}. We set channels sizes for the three stages to be ($C$, $2C$, $6C$).
  For the \textit{micro-level} designs, the employed MBConv-LN modifies MBConv~\cite{sandler2018mobilenetv2} by using a single LayerNorm~\cite{ba2016layer}. TFB-GeGLU upgrades TFB's FFNs~\cite{vaswani2017attention} (Feed-Forward Networks) with GELU Gated Linear Units~\cite{shazeer2020glu}.
  (b) In the CLIP framework, given $N$ image-text pairs, the  vision model's output $I_i$ is learned to align with its corresponding text Transformer's output $T_i$.
  Our text Transformers are the same as  OpenCLIP~\cite{ilharco_gabriel_2021_5143773}.
  +: Addition. *: Multiplication.
  }
  \label{fig:arch}
  \vspace{-3.5mm}
\end{figure*}

\vspace{-2mm}
\subsubsection{Macro-level Network Design} 
\label{sec:macro}
\vspace{-1mm}
\textbf{Overview:}
The macro-level network design of \modelname is inspired by the ViT and CoAtNet.
Specifically, on top of a simple convolutional stem (\ie, two $3\times3$ convolutions)~\cite{dai2021coatnet}, we adopt a 3-stage network architecture, where the first two stages employ the Mobile Convolution Blocks (MBConv)~\cite{sandler2018mobilenetv2,howard2019searching} and the third stage uses the Transformer Blocks (TFB)~\cite{vaswani2017attention,dosovitskiy2020image}.
\figref{fig:arch} shows the overview of \modelname.
We detail the design principles below, based on the discoveries from the re-benchmarking results

\textbf{Data and Model Scalability:}
ViT demonstrates the best scalability in terms of both model scales and data sizes. We thus opt for using Transformer Blocks in our last stage, and we stack most blocks here across different model sizes.

\textbf{Feature Resolution:}
We tailor the network to generate high resolution feature maps in the end.
Our 3-stage network design thus yields a feature map with output stride 16 (\ie, a downsampling factor of 16).

\textbf{Hybrid Architecture:}
Similar to CoAtNet, we employ MBConv in the first two stages, resulting in a hybrid model.
However, unlike CoAtNet that is constrained by its large memory usage, we propose a light-weight design of stage 1 and 2, which contain only two and four MBConv blocks.

Given the macro-level network design, we then move on to further improve the micro-level block design below.

\vspace{-2mm}
\subsubsection{Micro-level Block Design}
\label{sec:micro} 
\vspace{-1mm}
\textbf{Overview:}
The proposed \modelname depends on two types of blocks: Mobile Convolution Blocks (MBConv) and Transformer Blocks (TFB).
We refine each block in our model.

\textbf{MBConv-LN:}
The Mobile Convolution Block (MBConv)~\cite{sandler2018mobilenetv2} employs the ``inverted bottleneck'' design~\cite{he2016deep}, starting with a first $1\times1$ convolution to expand the channel size, followed by a $3\times3$ depthwise convolution~\cite{howard2017mobilenets} for spatial interaction, and ending with another $1\times1$ convolution to revert to the original channel size. Modern MBConv, as in MobileNetv3~\cite{howard2019searching}, adds numerous batch normalization (BN)~\cite{ioffe2015batch} layers and squeeze-and-excitation (SE)~\cite{hu2018squeeze}.
We adopt a simple modification by removing all BN layers and SE, and just using a single layer normalization (LN)~\cite{ba2016layer} as the first layer in our block, akin to the pre-norm layer in the Transformer block, resulting in the proposed MBConv-LN.
Ablation (in appendix) shows that MBConv-LN enjoys a simple design while attaining a similar performance to the original MBConv-BN-SE in MobileNetv3.

\textbf{TFB-GeGLU:}
The Transformer Block (TFB)~\cite{vaswani2017attention} contains two residual blocks: one with self-attention and the other with feed-forward network (FFN).
We empirically discover that substituting the first linear layer with GeGLU~\cite{shazeer2020glu}, an enhanced version of the Gated Linear Unit~\cite{dauphin2017language} that has a $2\times$ expansion rate, can enhance accuracy in the FFN.
We denote the Transformer Block with the updated FFN as TFB-GeGLU.
Ablation (in appendix) shows that TFB-GeGLU requires 12\% fewer parameters than TFB due to half expansion ratio, allowing us to stack additional Transformer blocks towards deeper architectures~\cite{szegedy2015going, zhou2021deepvit, touvron2021going}. 
\vspace{-3mm}

\subsubsection{Meta Architecture and Scaling Rule}
\label{sec:scale}

\textbf{Meta Architecture:}
After introducing our macro-level network and micro-level block designs, we now put everything together to form the meta architecture of \modelname.
Specifically, \modelname is a hybrid architecture that contains only three stages, built on top of a simple convolutional stem (\ie, two $3\times3$ convolutions).
The first two stages are composed of MBConv-LN, where we stack two and four of them for stage 1 and 2, respectively.
The third stage are obtained by stacking $N_B$ TFB-GeGLU blocks.
With the meta architecture in mind, we are ready to discuss the scaling rule to generate a family of \modelname with different model sizes.

\textbf{Scaling Rule:}
Our scaling rule is extremely simple and straightforward, controlled by two hyper-parameters: width (\ie, the channel sizes of those three stages) and depth (\ie, $N_B$, the number of TFB-GeGLU blocks in stage 3).
Note that our convolutional stem has the same channel size as the first stage.
We define four model sizes: Small, Base, Large, and X-Large (S, B, and L variants have a similar amount of model parameters to ViT~\cite{dosovitskiy2020image,zhai2022scaling}).
We use the same channel size as ViT in our 3rd stage for each model variant.
Specifically, we set the channel sizes of our three stages as $(C, 2C, 6C)$, where $6C = \{384, 768, 1024, 1152\}$ for Small, Base, Large and X-Large model variant, respectively\footnote{We calculate the channel size for stage 1 as $1/6C$, rounding to the nearest value that is divisible by 32.}.
Subsequently, given the target model parameter, the value of $N_B$ (\ie, the number of TFB-GeGLU blocks in stage 3) can be easily found.
We show the family of \modelname-\{S, B, L, XL\} in~\tabref{tab:family}.

\begin{table}[t!]
    \centering
    \scalebox{0.66}{
    \begin{tabular}{c c | cc | cc | cc | cc}
    &
    & \multicolumn{2}{c|}{\modelname-S} 
    & \multicolumn{2}{c|}{\modelname-B} 
    & \multicolumn{2}{c|}{\modelname-L} 
    & \multicolumn{2}{c}{\modelname-XL} 
    \\
    block & stride
    & \texttt{B} & \texttt{C} 
    & \texttt{B} & \texttt{C}  
    & \texttt{B} & \texttt{C}   
    & \texttt{B} & \texttt{C}   
    \\
    \shline
    conv-stem & 2
        & \texttt{2} & \texttt{64}
        & \texttt{2} & \texttt{128}
        & \texttt{2} & \texttt{160}
        & \texttt{2} & \texttt{192}
        \\
    MBConv-LN & 4
        & \texttt{2} & \texttt{64}
        & \texttt{2} & \texttt{128}
        & \texttt{2} & \texttt{160} 
        & \texttt{2} & \texttt{192} 
        \\
    MBConv-LN & 8
        & \texttt{4} & \texttt{128}
        & \texttt{4} & \texttt{256}
        & \texttt{4} & \texttt{320}
        & \texttt{4} & \texttt{384}
        \\
    TFB-GeGLU & 16
        & \texttt{14} & \texttt{384}
        & \texttt{14} & \texttt{768}
        & \texttt{31} & \texttt{1024} 
        & \texttt{32} & \texttt{1152} 
        \\
    \end{tabular}
    }
    \caption{
    \textbf{\modelname model variants.}
    \modelname variants 
    differ in the number of blocks \texttt{B} and number of channels \texttt{C} in each stage.
    }
    \label{tab:family}
    \vspace{-2mm}
\end{table}

\begin{table}[t!]
    \centering
    \scalebox{0.63}{
    \begin{tabular}{c | ccc | ccc}
    & \multicolumn{3}{c|}{\textbf{short schedule} for benchmarking} 
    & \multicolumn{3}{c}{\textbf{long schedule}}
    \\
    & \modelname-S & \modelname-B & \modelname-L & \modelname-L & \multicolumn{2}{c}{\modelname-XL}  
    \\
    \shline
    batch size & 8k & 8k & 16k & 90k & 90k & 90k\\
    image size & 224 & 224 & 224 & 224 & 256 & 256 \\
    \# A100 GPUs & 32 & 32 & 32 & 184 & 312 & 312\\
    \# epochs & 1 & 1 & 1 & 10 & 10 & 30 \\
    seen samples (B) & 1.28 & 1.28 & 1.28 & 12.8 & 12.8 & 40.0 \\
    training days & 1.8 & 3.3 & 5.6 & 11 & 15 & 46 \\
    \end{tabular}
    }
    \caption{
    \textbf{Short and long training schedules on DataComp-1B.}
    }
    \label{tab:training_protocol}
    \vspace{-5mm}
\end{table}

\textbf{Locked-Text Tuning for CLIP:}
\label{sec:ltt}
Besides model design, we propose Locked-Text Tuning (LTT) to exploits a pretrained frozen text encoder.
In light of the aligned image and text embeddings in CLIP, we leverage the pretrained text encoder from a large VLM to guide the training of image encoders of smaller VLMs.
Specifically, when training other \modelname variants (\eg, \modelname-S and \modelname-B), we initialize their text encoder with the one from a pretrained \modelname-L.
The text encoder is then frozen, used as a teacher to guide the training of the randomly initialized image encoder. 
This scheme can be considered as a way to distill the knowledge~\cite{hinton2015distilling} from a pretrained frozen text encoder to a randomly initialized image encoder.

\begin{table*}[ht]
\centering
\resizebox{0.95\textwidth}{!}{
\begin{tabular}{c|c|c|c|c|c|c|c|c|c|c|c|c|c}
 & image  & num & text encoder & seen & training & training & trainable params & MACs   & ImageNet & avg. 38 &   ImageNet & VTAB & retrieval \\
image encoder & size & patches & depth/width  & samples (B) & scheme  & dataset & Image+Text (M) & Image+Text (G)  & Acc. & datasets &  dist. shift. &  &  \\
\shline
 ViT-L/14~\cite{gadre2023datacomp} & 224 & 256 & 12 / 768 & 12.8 & OpenCLIP & DataComp-1B & 304.0 + 123.7 & 77.8 + 6.6 & 79.2 & 66.3 & 67.9 & 65.2 & 60.8 \\
  ViT-L/14~\cite{li2023clipa} & 224 & 256 & 12 / 768 & 12.8 + 0.5 & CLIPA-v2 & DataComp-1B & 304.0 + 110.3 & 77.8 + 2.7 & 79.7 & 65.4 & 68.6 & 62.9 & 60.6 \\
    ViT-L/14~\cite{li2023clipa} & 336 & 576 & 12 / 768 & 12.8 + 0.5 + 0.1 & CLIPA-v2 & DataComp-1B & 304.3 + 110.3 & 174.7 + 2.7 & 80.3 & 65.7 & 70.2 & 62.5 & 61.1 \\
\hline \hline
  \modelname-L & 224 & 196 & 12 / 768 & 12.8 & OpenCLIP & DataComp-1B & 333.3 + 123.7 & 72.6 + 6.6 & 80.8 & 66.7 & 69.8 & 65.3 & 60.3 \\
   \modelname-L & 256$^\dagger$ & 256 & 12 / 768 & 12.8 + 0.2 & OpenCLIP & DataComp-1B & 333.4 + 123.7 & 94.8 + 6.6 & 81.2 & 67.0 & 71.1 & 65.3 & 61.2 \\
  \modelname-L & 336 & 441 & 12 / 768 & 12.8 + 0.2 & OpenCLIP & DataComp-1B & 333.6 + 123.7 & 163.4 + 6.6 & 81.6  & 67.0 & 72.1 & 64.4 & 61.6 \\
  \modelname-L & 384$^\dagger$ & 576 & 12 / 768 & 12.8 + 0.2 & OpenCLIP & DataComp-1B & 333.7 + 123.7 & 213.4 + 6.6 & 81.8 & 67.2 & 72.4 & 64.7 & 61.8 \\
\hline
  \modelname-L2 & 224 & 196 & 24 / 1024 & 12.8 & OpenCLIP & DataComp-1B & 333.6 + 354.0 & 72.6 + 23.3 & 80.9 & 66.4 & 70.6 & 63.4 & 61.5 \\
   \modelname-L2 & 256$^\dagger$ & 256 & 24 / 1024 & 12.8 + 0.5 & OpenCLIP & DataComp-1B & 333.6 + 354.0 & 94.8 + 23.3 & 81.5 & 67.4 & 71.9 & 64.1 & 63.1 \\
  \modelname-L2 & 336 & 441 & 24 / 1024 & 12.8 + 0.5 & OpenCLIP & DataComp-1B & 333.8 + 354.0 & 163.4 + 23.3 & 81.8 & 67.8 & 73.0 & 64.5 & 63.6 \\
  \modelname-L2 & 384$^\dagger$ & 576 & 24 / 1024 & 12.8 + 0.5 & OpenCLIP & DataComp-1B & 334.0 + 354.0 & 213.4 + 23.3  & 82.1 & 68.1 & 73.4 & 64.8 & 63.7 \\
  \hline
  \modelname-XL & 256$^\dagger$ & 256 & 27 / 1152 & 12.8 + 0.5 & OpenCLIP & DataComp-1B & 436.1 + 488.7 & 125.3 + 33.1 & 82.1 & 67.6 & 72.3 & 65.4 & 62.7 \\
  \modelname-XL & 384$^\dagger$ & 576 & 27 / 1152 & 12.8 + 0.5 & OpenCLIP & DataComp-1B & 436.1 + 488.7 & 281.9 + 33.1 & 82.6 & 68.1 & 73.6 & \textbf{65.6} & \textbf{63.8} \\
  \hline
    \modelname-XL & 256$^\dagger$ & 256 & 27 / 1152 & 40.0 & OpenCLIP & DataComp-1B & 436.1 + 488.7 & 125.3 + 33.1 & 82.3 & 67.5 & 72.8 & 64.0 & 62.1 \\
  \modelname-XL & 336$^\dagger$ & 441 & 27 / 1152 & 40.0 + 1.0 & OpenCLIP & DataComp-1B & 436.1 + 488.7 & 215.9 + 33.1 & 82.7 & 68.0 & 73.9 & 64.1 & 62.6 \\
  \modelname-XL & 384$^\dagger$ & 576 & 27 / 1152 & 40.0 + 1.0 & OpenCLIP & DataComp-1B & 436.1 + 488.7 & 281.9 + 33.1 & \textbf{82.9} & \textbf{68.1} & \textbf{74.1} & 64.0 & 62.5 \\

\hline \hline
    \textcolor{siglipcolor}{ViT-L/14}~\cite{sun2023eva} & \textcolor{siglipcolor}{224} & \textcolor{siglipcolor}{256} & \textcolor{siglipcolor}{12 / 768} & \textcolor{siglipcolor}{4.0} & \textcolor{siglipcolor}{EVA-CLIP} & \textcolor{siglipcolor}{Merged-2B} & \textcolor{siglipcolor}{333.3 + 123.7} & \textcolor{siglipcolor}{72.6 + 6.6} & \textcolor{siglipcolor}{79.8} & \textcolor{siglipcolor}{64.9} & \textcolor{siglipcolor}{68.9} & \textcolor{siglipcolor}{62.8} & \textcolor{siglipcolor}{63.3} \\
    \textcolor{siglipcolor}{ViT-L/14}~\cite{sun2023eva} & \textcolor{siglipcolor}{336} & \textcolor{siglipcolor}{576} & \textcolor{siglipcolor}{12 / 768} & \textcolor{siglipcolor}{6.0} & \textcolor{siglipcolor}{EVA-CLIP} & \textcolor{siglipcolor}{Merged-2B} & \textcolor{siglipcolor}{333.3 + 123.7} & \textcolor{siglipcolor}{72.6 + 6.6} & \textcolor{siglipcolor}{80.4} & \textcolor{siglipcolor}{65.8} & \textcolor{siglipcolor}{70.9} & \textcolor{siglipcolor}{63.2} & \textcolor{siglipcolor}{63.5} \\
  \textcolor{siglipcolor}{ViT-L/16}~\cite{zhai2023sigmoid} & \textcolor{siglipcolor}{256} & \textcolor{siglipcolor}{256} & \textcolor{siglipcolor}{24 / 1024} & \textcolor{siglipcolor}{40.0} & \textcolor{siglipcolor}{SigLIP} & \textcolor{siglipcolor}{WebLI} & \textcolor{siglipcolor}{316.0 + 336.2} & \textcolor{siglipcolor}{78.1 + 19.3}  & \textcolor{siglipcolor}{80.5} & \textcolor{siglipcolor}{65.6} & \textcolor{siglipcolor}{70.2} & \textcolor{siglipcolor}{62.5} & \textcolor{siglipcolor}{61.1} \\
  \textcolor{siglipcolor}{ViT-L/16}~\cite{zhai2023sigmoid} & \textcolor{siglipcolor}{384} & \textcolor{siglipcolor}{576} &  \textcolor{siglipcolor}{24 / 1024} & \textcolor{siglipcolor}{40.0 + 5.0} & \textcolor{siglipcolor}{SigLIP} & \textcolor{siglipcolor}{WebLI} & \textcolor{siglipcolor}{316.3 + 336.2} & \textcolor{siglipcolor}{175.8 + 19.3}  & \textcolor{siglipcolor}{82.1} & \textcolor{siglipcolor}{66.8} & \textcolor{siglipcolor}{70.9} & \textcolor{siglipcolor}{63.1} & \textcolor{siglipcolor}{68.7} \\
  \hline 
\textcolor{siglipcolor}{ViT-G/14}~\cite{ilharco_gabriel_2021_5143773} & \textcolor{siglipcolor}{224} & \textcolor{siglipcolor}{256} & \textcolor{siglipcolor}{32 / 1280} & \textcolor{siglipcolor}{39.0}  & \textcolor{siglipcolor}{OpenCLIP} & \textcolor{siglipcolor}{LAION-2B} & \textcolor{siglipcolor}{1844.9 + 694.7} & \textcolor{siglipcolor}{473.4 + 48.5} & \textcolor{siglipcolor}{80.1} & \textcolor{siglipcolor}{66.7} & \textcolor{siglipcolor}{69.1} & \textcolor{siglipcolor}{64.6} & \textcolor{siglipcolor}{63.5} \\
 ViT-H/14~\cite{li2023clipa} & 336 & 576 & 24 / 1024 & 12.8 + 0.5 + 0.1 & CLIPA-v2 & DataComp-1B & 632.5 + 354.0 & 363.7 + 9.7  & 81.8 & 66.8 & 72.4 & 63.7 & 62.6 \\
 \textcolor{siglipcolor}{ViT-E/14}~\cite{sun2023eva} & \textcolor{siglipcolor}{224} & \textcolor{siglipcolor}{256} & \textcolor{siglipcolor}{24 / 1024} & \textcolor{siglipcolor}{4.0}  & \textcolor{siglipcolor}{EVA-CLIP} & \textcolor{siglipcolor}{LAION-2B} & \textcolor{siglipcolor}{4350.6 + 354.0} & \textcolor{siglipcolor}{1117.3 + 23.3} & \textcolor{siglipcolor}{82.0} & \textcolor{siglipcolor}{66.9} & \textcolor{siglipcolor}{72.0} & \textcolor{siglipcolor}{63.6} & \textcolor{siglipcolor}{62.8} \\
  ViT-G/14~\cite{li2023clipa} & 336 & 576 & 32 / 1280 & 12.8 + 0.5 + 0.1  & CLIPA-v2 & DataComp-1B & 1845.4 + 672.3 & 1062.9 + 20.2 & 83.1 & 68.4 & 74.0 & 64.5 & 63.1 \\
   \textcolor{siglipcolor}{SoViT-400M/14}~\cite{alabdulmohsin2023getting} & \textcolor{siglipcolor}{224} & \textcolor{siglipcolor}{256} & \textcolor{siglipcolor}{27 / 1152} & \textcolor{siglipcolor}{40.0} & \textcolor{siglipcolor}{SigLIP} & \textcolor{siglipcolor}{WebLI} & \textcolor{siglipcolor}{428.2 + 449.7} & \textcolor{siglipcolor}{106.2 + 6.6}  & \textcolor{siglipcolor}{82.0} & \textcolor{siglipcolor}{68.1} & \textcolor{siglipcolor}{69.5} & \textcolor{siglipcolor}{64.8} & \textcolor{siglipcolor}{66.8} \\ 
   \textcolor{siglipcolor}{SoViT-400M/14}~\cite{alabdulmohsin2023getting} & \textcolor{siglipcolor}{384} & \textcolor{siglipcolor}{729} & \textcolor{siglipcolor}{27 / 1152} & \textcolor{siglipcolor}{40.0 + 5.0} & \textcolor{siglipcolor}{SigLIP} & \textcolor{siglipcolor}{WebLI} & \textcolor{siglipcolor}{428.2 + 449.7} & \textcolor{siglipcolor}{302.3 + 26.3}  & \textcolor{siglipcolor}{83.1} & \textcolor{siglipcolor}{69.2} & \textcolor{siglipcolor}{72.4} & \textcolor{siglipcolor}{64.6} & \textcolor{siglipcolor}{69.8} \\
   \textcolor{siglipcolor}{ViT-H/14}~\cite{fang2023data} & \textcolor{siglipcolor}{224} & \textcolor{siglipcolor}{256} & \textcolor{siglipcolor}{24 / 1024} & \textcolor{siglipcolor}{39.0} & \textcolor{siglipcolor}{OpenCLIP} & \textcolor{siglipcolor}{DFN-5B} & \textcolor{siglipcolor}{632.1 + 354.0} & \textcolor{siglipcolor}{162.0 + 23.3}   & \textcolor{siglipcolor}{83.4} & \textcolor{siglipcolor}{69.6} & \textcolor{siglipcolor}{69.9} & \textcolor{siglipcolor}{67.5} & \textcolor{siglipcolor}{68.3} \\
   \textcolor{siglipcolor}{ViT-H/14}~\cite{fang2023data} & \textcolor{siglipcolor}{378} & \textcolor{siglipcolor}{729} & \textcolor{siglipcolor}{24 / 1024} & \textcolor{siglipcolor}{39.0 + 5.0} & \textcolor{siglipcolor}{OpenCLIP} & \textcolor{siglipcolor}{DFN-5B} & \textcolor{siglipcolor}{632.7 + 354.0} & \textcolor{siglipcolor}{460.1 + 23.3}  & \textcolor{siglipcolor}{84.4} & \textcolor{siglipcolor}{70.8} & \textcolor{siglipcolor}{72.8} & \textcolor{siglipcolor}{68.5} & \textcolor{siglipcolor}{69.5}
\end{tabular}
 }
\caption{
\textbf{Comparison with state-of-the-art models.}
Our models are only trained on the publicly available DataComp-1B~\cite{gadre2023datacomp}. CLIPA-v2~\cite{li2023clipa} uses an advanced progressive training scheme (from smaller images to larger ones) than the original OpenCLIP~\cite{ilharco_gabriel_2021_5143773, gadre2023datacomp} scheme that we follow.
Other methods that use different settings are marked in gray for reference.
Specifically, EVA-CLIP~\cite{sun2023eva} uses EVA weights~\cite{fang2023eva}, better training scheme FLIP~\cite{li2023scaling}, and different training datasets~\cite{schuhmann2022laion,fang2023eva}.
SigLIP~\cite{zhai2023sigmoid} employs better sigmoid loss, stronger text encoders, and an extremely long schedule on the proprietary WebLI dataset~\cite{chen2022pali} (40B for training and another 5B seen samples for fine-tuning).
$\dagger$: ViT-L/14 benefits from more image tokens by using a smaller output stride 14 than 16 that we use. To have the same image tokens, we slightly enlarge the image size (\eg, $224/14 = 256/16$ and $336/14 = 384/16$). We note that all compared results are from the \href{https://github.com/mlfoundations/open_clip/blob/main/docs/openclip_results.csv}{OpenCLIP-results} that are evaluated under the same setting to ensure a fair comparison.
}
\label{tab:sota}
\vspace{-5mm}
\end{table*}

%% file: sec/5_experiment.tex
\section{Experimental Results}
\label{sec:results}

In this section, we detail the implementation in Sec.~\ref{sec:imp},  compare with the state-of-the-arts in Sec.~\ref{sec:sota}, and deploy \modelname to downstream tasks, including open-vocabulary detection/segmentation, and large multi-modal models in Sec.~\ref{sec:downstream}. See appendix for ablation studies.

\subsection{Implementation Details} 
\label{sec:imp}

\textbf{Training Strategy:}
We train the VLMs using OpenCLIP~\cite{ilharco_gabriel_2021_5143773} on the public dataset DataComp-1B~\cite{gadre2023datacomp}. 
\tabref{tab:training_protocol} summarizes the settings for our training schedules and model variants. We use the short schedule to benchmark vision models and conduct our ablation studies, and long schedule to train our best \modelname-L.
We closely follow the training hyper-parameter settings in OpenCLIP~\cite{ilharco_gabriel_2021_5143773,gadre2023datacomp}. The training and fine-tuning details are in the appendix. 


\textbf{Evaluation Strategy:} We follow DataComp~\cite{gadre2023datacomp} to zero-shot evaluate VLMs with a testbed of 38 tasks, including ImageNet~\cite{russakovsky2015imagenet}, 6 distribution shift tasks~\cite{wang2019learning,barbu2019objectnet,recht2019imagenet,hendrycks2021natural,hendrycks2021many}, VTAB tasks~\cite{zhai2019visual}, WILDS tasks~\cite{koh2021wilds,sagawa2022extending}, and 3 retrieval tasks~\cite{young2014image,chen2015microsoft,bitton2022winogavil}.

\textbf{Other Downstream Tasks:}
We evaluate the trained VLM in downstream tasks. For open-vocabulary detection, we exploit the F-ViT framework~\cite{wu2023clipself}, while for open-vocabulary segmentation, we adopt the FC-CLIP framework~\cite{yu2023convolutions} and zero-shot evaluate on multiple segmentation datasets.
Finally, we evaluate VLMs in LLaVA-1.5~\cite{liu2023improvedllava} for LMMs across multiple benchmarks.
In all the cases, F-ViT, FC-CLIP, and LLaVA employ the frozen VLM backbone to effectively ablate different pretrained VLMs.


\subsection{Main Results}
\label{sec:sota}

\textbf{Comparison with other State-of-the-arts:}
\tabref{tab:sota} summarizes the comparison between \modelname-L and other state-of-the-art models, which exclusively employ the ViT backbone~\cite{dosovitskiy2020image} but use different training schemes and datasets.
For a fair comparison, we focus on the methods that use the same training data DataComp-1B~\cite{gadre2023datacomp}, but still list other methods in the table for reference.
For simplicity, we use ``X@Z'' to denote the vision model X trained with input size Z
\footnote{Notation @ here is slightly abused to denote the training seen samples.}.
ImageNet zero-shot accuracy is our main metric; other results are still reported in the table.
As shown in the table, \modelname-L@224 outperforms ViT-L/14@224 OpenCLIP~\cite{ilharco_gabriel_2021_5143773} by +1.6\%.
However, ViT-L/14 benefits from more image tokens by using a smaller output stride 14 than 16 that we use (as benchmarked in the appendix).
To have the same image tokens, we slightly enlarge the image size.
As a result, our \modelname-L@256 surpasses ViT-L/14@224 OpenCLIP~\cite{ilharco_gabriel_2021_5143773} and CLIPA-v2~\cite{li2023clipa} by 2.0\% and 1.5\%, respectively.
After fine-tuning on larger input sizes, \modelname-L@384 and \modelname-L@336 still performs better than ViT-L/14@336 CLIPA-v2~\cite{li2023clipa} by +1.5\% and +1.3\%, respectively. 
Impressively, with only half the parameters, our \modelname-L attains an average of 67.2\% performance across 38 datasets, exceeding the larger ViT-H/14 CLIPA-v2 model's performance by +0.4\%. Scaling up the text encoder to match the model size of the image encoder (specifically, \modelname-L2) notably increases zero-shot ImageNet accuracy to 82.1\% and average 38 datasets performance to 68.1\%.
Further scaling up the model parameters (\ie, \modelname-XL) and 40 billion seen samples reaches 82.9\% zero-shot ImageNet accuracy.

\begin{figure}
  \centering
  \includegraphics[width=0.8\linewidth]{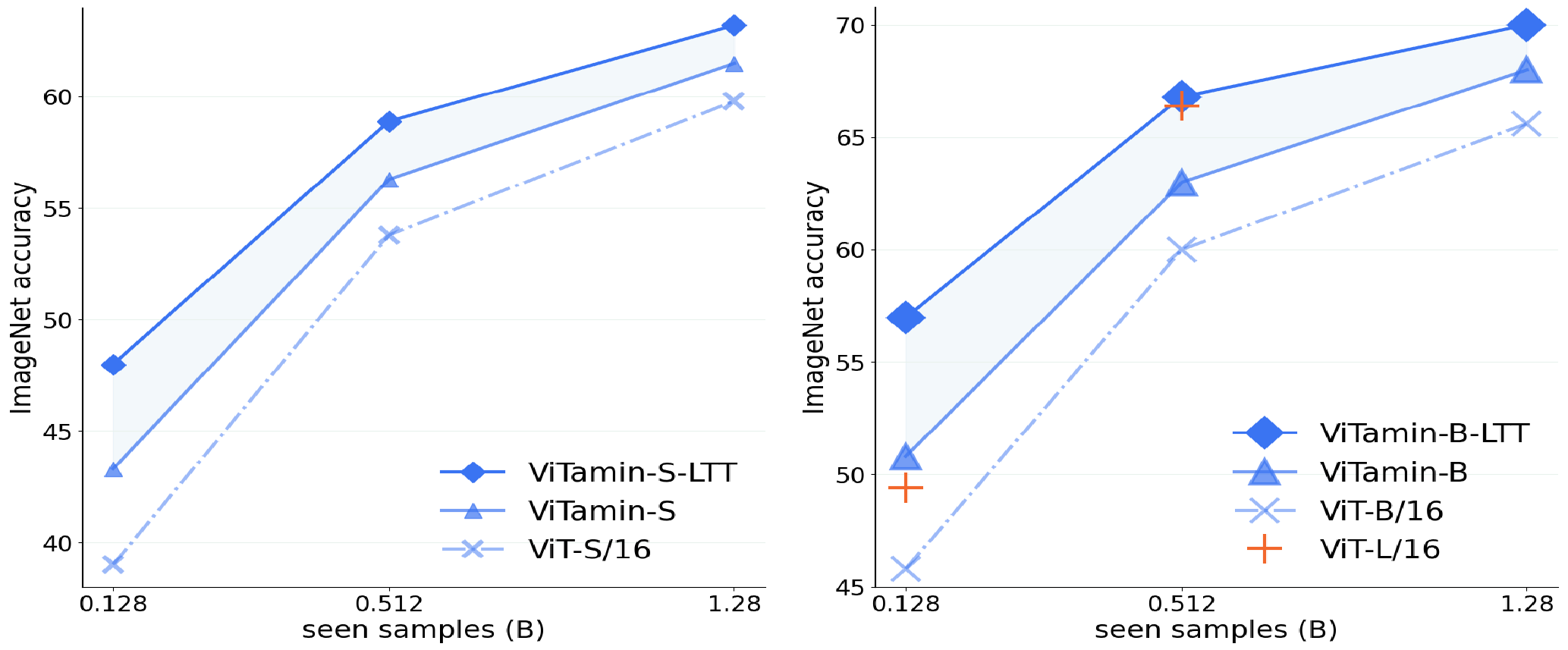}
  \vspace{-1mm}
  \caption{\textbf{Locked-text tuning (LTT)}. LTT exploits a pretrained frozen text encoder, and effectively boosts the model performance.
  }
  \label{fig:ltt}
  \vspace{-4mm}
\end{figure}

\textbf{Locked-Text Tuning:} 
\figref{fig:ltt} shows that our LTT improves our \modelname-S/-B by a large margin, especially when data sizes are small. Notably, LTT lifts \modelname-B to the next scale of model performance, surpassing ViT-L/16 by +14\% in 128M samples and +1.1\% in 512M seen samples. Interestingly, LTT can save 10\% training budget for \modelname-B as the text tower is completely frozen.

\textbf{Data Quality \vs Model Capacity:}
The DataComp challenge~\cite{gadre2023datacomp} underscores the role of data filtering for VLM, however, using a fixed ViT model. 
As shown in~\tabref{tab:data_quality}, the leading solution~\cite{yu2023devil} of DataComp challenge in ICCV 2023 employed a complicated 24 filtering rules to improve the dataset quality, resulting in +2.3\% gain. 
Surprisingly, our \modelname-B improves the performance by a healthy margin of +12.8\% accuracy, and locked-text tuning can lift the gain to +23.3\%. The result highlights the importance to co-design the vision-language dataset and model.

\begin{table}[t!]
    \centering
    \resizebox{0.93\columnwidth}{!}{
    \begin{tabular}{c|c|c|c|c|c}

       & & dataset  & seen  & IN acc. & avg. 38 \\
         image encoder & data filtering &  size &   samp. & (\%) &  datasets (\%) \\
        \shline
        \multicolumn{6}{l}{\baseline{\textit{\textbf{leaderboard}}}} \\ 
        ViT-B/32 & DataComp~\cite{gadre2023datacomp} & 14M & 128M & 29.7 & 32.8 \\
        ViT-B/32 & SIEVE~\cite{mahmoud2023sieve} & 24M & 128M & 30.3	\textcolor{blue}{(+0.6)} & 35.3 \textcolor{blue}{(+2.5)}\\
        ViT-B/32 & Top-1 Solution~\cite{yu2023devil} & 23M  & 128M & 32.0 \textcolor{blue}{(+2.3)}	 & 37.1 \textcolor{blue}{(+4.3)}\\

        \shline
        \multicolumn{6}{l}{\baseline{\textit{\textbf{our experiments}}}} \\ 
        ViT-B/32 &  DataComp~\cite{gadre2023datacomp} & 14M & 128M & 29.4 & 31.5 \\
        ViT-B/16 & DataComp~\cite{gadre2023datacomp} & 14M & 128M & 35.8 \textcolor{blue}{(+6.4)}& 34.6 \textcolor{blue}{(+3.1)}\\
        \modelname-B & DataComp~\cite{gadre2023datacomp} & 14M & 128M & 42.2 \textcolor{blue}{(+12.8)}& 38.3 \textcolor{blue}{(+6.8)}\\
        \hline
        ViT-B/16-LTT & DataComp~\cite{gadre2023datacomp} & 14M & 128M & 43.6 \textcolor{blue}{(+14.2)}& 41.1 \textcolor{blue}{(+9.6)}\\
        \modelname-B-LTT & DataComp~\cite{gadre2023datacomp} & 14M & 128M & 52.7 \textcolor{blue}{(+23.3)} & 47.2 \textcolor{blue}{(+15.7)}\\

    \end{tabular}
    }
    \vspace{-1.5mm}
    \caption{
    \textbf{Data quality \vs model capacity.} The leaderboard results are from ICCV 2023 DataComp challenge medium filtering track.
    }
    \vspace{-1.5mm}
    \label{tab:data_quality}
\end{table}

\subsection{New Suite of Downstream Tasks}
\label{sec:downstream}
The evaluations so far are mostly on classification/retrieval-based task, highlighting a lack of downstream tasks similar to those employed in the ImageNet era. Yet, in contrast to ImageNet-based vision models where downstream tasks mainly involve transfer learning for conventional detection and segmentation, VLMs excel with zero-shot capability and provides feature embeddings that are well-aligned across the vision-language domain. In light of this, we introduce a novel suite of downstream tasks aimed at the holistic evaluation of VLMs, including open-vocabulary detection and segmentation and multi-modal LLM.

\begin{table}[!t]
\small
\centering
\scalebox{0.78}{
\tablestyle{4pt}{1.05}
\begin{tabular}{c|cc|cc}
 &  \multicolumn{2}{c|}{pretraining}  & OV-COCO \cite{zareian2021open} & OV-LVIS~\cite{gu2021open}  \\
image encoder & dataset & scheme &  ($\text{AP}^{\text{novel}}_{\text{50}})$ & ($\text{AP}_{\text{r}}$) \\
\shline
ViT-L/14 & DataComp-1B & CLIPA-v2   & 36.1 & 32.5 \\
\textcolor{siglipcolor}{ConvNeXt-L} & \textcolor{siglipcolor}{LAION-2B} & \textcolor{siglipcolor}{OpenCLIP}  & \textcolor{siglipcolor}{36.4} & \textcolor{siglipcolor}{29.1}  \\
\hline 
\modelname-L & DataComp-1B & OpenCLIP   & 37.5 & 35.6  \\
\end{tabular}
}
\vspace{-1mm}
\caption{
    \label{tab:downstream_openvocabdet}
    \textbf{Open-vocabulary detection.}
    Different image encoders (ViT-L/14 by~\cite{li2023clipa}, ConvNeXt-L by~\cite{ilharco_gabriel_2021_5143773}) are using the F-ViT framework~\cite{wu2023clipself} in a sliding window manner~\cite{yu2023towards}, trained on OV-COCO~\cite{zareian2021open} and OV-LVIS~\cite{gu2021open}. ConvNeXt-L is marked in gray due to different pretrained dataset.
}
\vspace{-1.5mm}
\label{tab:open-voc-det}
\end{table}

\begin{table}[!t]
\small
\centering
\scalebox{0.69}{
\tablestyle{2pt}{1.1}
\begin{tabular}{c|cc|ccc|cccccc}
                           &  & & \multicolumn{3}{c|}{panoptic dataset (PQ)}   & \multicolumn{5}{c}{semantic dataset (mIoU)}                                \\
image  &  \multicolumn{2}{c|}{pretraining}  & ADE            & Cityscapes   & MV   & A-150 &A-847 & PC-459 & PC-59 & PAS-21             \\
 encoder & dataset & scheme & \cite{zhou2017scene} & \cite{cordts2016cityscapes} & \cite{neuhold2017mapillary} &  \cite{zhou2017scene} & \cite{zhou2017scene} & \cite{mottaghi2014role} & \cite{mottaghi2014role} & \cite{everingham2010pascal}\\
\shline
ViT-L/14 & DataComp-1B & CLIPA-v2   & 24.6 & 40.7 & 16.5 & 31.8 & 14.3 & 18.3 & 55.1 & 81.5 \\
\textcolor{siglipcolor}{ConvNeXt-L} & \textcolor{siglipcolor}{LAION-2B} & \textcolor{siglipcolor}{OpenCLIP}  & \textcolor{siglipcolor}{26.8} & \textcolor{siglipcolor}{\textbf{44.0}} & \textcolor{siglipcolor}{\textbf{18.3}} & \textcolor{siglipcolor}{34.1} & \textcolor{siglipcolor}{14.8} & \textcolor{siglipcolor}{18.2} & \textcolor{siglipcolor}{\textbf{58.4}} & \textcolor{siglipcolor}{81.8} \\
\hline 
\modelname-L & DataComp-1B & OpenCLIP   & 27.3 & 44.0 & 18.2 & 35.6 & 16.1 & 20.4 & 58.4 & 83.4 \\
\end{tabular}
}
\caption{
    \label{tab:downstream_openvocabseg}
    \textbf{Open-vocabulary segmentation.}
    Different image encoders (ViT-L/14 by~\cite{li2023clipa}, ConvNeXt-L by~\cite{ilharco_gabriel_2021_5143773}) are using the FC-CLIP framework~\cite{yu2023convolutions} in a sliding window manner~\cite{yu2023towards}, trained on COCO~\cite{lin2014microsoft} and zero-shot evaluated on the other datasets. ConvNeXt-L is marked in gray due to different pretrained dataset.
}
\label{tab:open-voc}
\vspace{-2.5mm}
\end{table}

\textbf{Open-Vocabulary Detection and Segmentation:} To examine how well the trained VLMs can adapt to downstream tasks, we consider two simple yet effective frameworks F-ViT~\cite{wu2023clipself} and FC-CLIP~\cite{yu2023convolutions} which utilize a frozen CLIP backbone for open-vocabulary detection and segmentation, respectively.  
Specifically, we consider different VLMs as plug-in frozen backbones to these frameworks, while for ViT and \modelname that may not easily generalize to high resolution input, we extract the feature in a sliding window manner~\cite{yu2023towards}, with window size equal to the pre-train image size, resulting in \ovdetmodelname and \ovsegmodelname, respectively.
\tabref{tab:downstream_openvocabdet} illustrates that \modelname-L serves as a stronger image encoder for open-vocabulary detection, surpassing its ViT-L/14 counterpart by 1.4\% and 3.1\% on OV-COCO and OV-LVIS. \tabref{tab:downstream_openvocabseg} shows that \modelname-L outperforms ViT-L/14 by 2.6\% on average 3 panoptic datasets and by 2.6\% on average 5 semantic datasets.  Notably, surpassing prior art, \modelname-L sets a new state-of-the-art performance across seven benchmarks for open-vocabulary panoptic segmentation and semantic segmentation. 


\textbf{Large Multi-modal Models:} Another key application of VLMs lies in their role as vision encoders within LMMs ~\cite{li2022blip,zhu2023minigpt,liu2023visual}, as image features in VLMs that is well-aligned with text, thereby bridging the visual comprehension gap for LLMs. Specifically, we consider LLaVA-1.5~\cite{liu2023improvedllava} as the evaluated framework. 
We follow~\cite{liu2023improvedllava} for all experimental settings, where the image is processed through a frozen CLIP model and a MLP projector, retaining the image as visual tokens, which are prepended to a text sequence and fed into a frozen Vicuna-v1.5-7B~\cite{vicuna2023}. We run evaluation on 12 LMM benchmarks following~\cite{liu2023improvedllava}, with results in \tabref{tab:downstream_llava}. It should be noted that while OpenAI-trained ViT-L/14 underperforms CLIPAv2-trained counterpart by -3.7\% ImageNet accuracy, it excels remarkably in LLaVA (+4.4\% on VQAv2 and +4.3\% on VizWiz). This highlights the need for incorporating a variety of downstream tasks to ensure a comprehensive evaluation. Surprisingly, simply replacing LLaVA's image encoder to \modelname-L can achieve new state-of-the-art across various benchmarks.

\begin{table}[!t]
\small
\centering
\scalebox{0.78}{
\tablestyle{2pt}{1.1}
\begin{tabular}{cc|ccccccccccccc}
image      & training             & \rotatebox{90}{VQAv2}            & \rotatebox{90}{GQA}   & \rotatebox{90}{VizWiz}   & \rotatebox{90}{SQA} & \rotatebox{90}{T-VQA} & \rotatebox{90}{POPE} & \rotatebox{90}{MME} & \rotatebox{90}{MMBench} & \rotatebox{90}{MMB$^{CN}$} & \rotatebox{90}{SEED} & \rotatebox{90}{LLaVA$^W$} & \rotatebox{90}{MM-Vet}             \\
 encoder     & scheme   & \cite{goyal2017making}          & \cite{hudson2019gqa}   & \cite{gurari2018vizwiz}   & \cite{lu2022learn} & \cite{singh2019towards} & \cite{fu2023mme} & \cite{li2023evaluating} & \cite{liu2023mmbench} & \cite{liu2023mmbench} & \cite{li2023seed} & \cite{liu2023visual} &  \cite{yu2023mm}            \\

\shline
\textcolor{siglipcolor}{ViT-L/14}  & \textcolor{siglipcolor}{OpenAI} & \textcolor{siglipcolor}{78.5} & \textcolor{siglipcolor}{\textbf{62.0}} & \textcolor{siglipcolor}{50.0} & \textcolor{siglipcolor}{66.8} & \textcolor{siglipcolor}{58.2} & \textcolor{siglipcolor}{\textbf{85.9}} & \textcolor{siglipcolor}{\textbf{1511}} & \textcolor{siglipcolor}{64.3} & \textcolor{siglipcolor}{58.3} & \textcolor{siglipcolor}{\textbf{58.6}} & \textcolor{siglipcolor}{65.4} & \textcolor{siglipcolor}{31.1} \\
\hline \hline
ViT-L/14  & CLIPA-v2 &  75.9 & 60.3 & 48.8 & 65.6 & 55.0 & 84.9 & 1396 & 60.8 & 54.6 & 54.6 & 60.6 & 28.6 \\
\hline
\modelname-L   & OpenCLIP & 78.4 & 61.6 & 51.1 & 66.9 & 58.7 & 84.6 & 1421 & \textbf{65.4} & \textbf{58.4} & 57.7 & 64.5 & \textbf{33.6} \\
\modelname-L$^\dagger$ & OpenCLIP & \textbf{78.9} & 61.6 & \textbf{55.4} & \textbf{67.6} & \textbf{59.8} & 85.5 & 1447 & 64.5 & 58.3 & 57.9 & \textbf{66.1} & \textbf{33.6} \\
\end{tabular}
}
\caption{
    \textbf{Large Multi-modal Model (LMM) performance with different VLMs.} The results in 1st row originate from LLaVa-1.5 paper~\cite{liu2023improvedllava} and are marked in gray due to pretraining on OpenAI WIT dataset~\cite{radford2021learning} unlike DataComp-1B~\cite{gadre2023datacomp} used by other rows.  
    All listed models are trsained following the same settings in LLaVA-1.5~\cite{liu2023improvedllava} with Vicuna-V1.5-7B~\cite{vicuna2023}, for a fair comparison. $\dagger$:  image size of 384 rather than the default 336. 
}
\label{tab:downstream_llava}
\vspace{-4mm}
\end{table}

%% file: sec/6_conclusion.tex
\section{Conclusion}
In this work, we build an evaluation protocols of modern vision models in VLM and re-benchmark them under CLIP setting. We examine vision models from four aspects of data scalability, model scalability, feature resolution and hybrid architecture. The four pillars motivate us to propose \modelname, which not only competes favorably with ViT in zero-shot ImageNet accuracy and average 38 dataset accuracy, but also achieves the state-of-the-art on 22 downstream tasks covering open-vocaburary detection and segmentation and large multi-modal models. 
We hope that our design practices will drive the development of more advanced vision models for VLMs.

\textbf{Acknowledgement}: We thank Haichao Yu and Zhanpeng Zeng for the discussions about DataComp challenge and micro-level block design, respectively.
This work was supported in part by ONR N00014-23-1-2641.

%% file: sec/X_suppl.tex
\appendix
\section*{Appendix}
\label{sec:appendix}
In the supplementary materials, we provide additional information, as listed below.

\begin{itemize}
    \item Sec.~\ref{sec:abl}: The ablation studies on the \modelname macro-level network and micro-level block designs.
     \item Sec.~\ref{sec:ovlvis}: \modelname sets new SoTA in open-vocabulary dense prediction tasks including the OV-LVIS detection benchmark and 6 segmentation benchmarks.
    \item Sec.~\ref{sec:appendix_ltt}: The results of using the proposed Locked-Text Tuning (LTT) training scheme.
    \item Sec.~\ref{sec:imagenet22k}: The results of benchmarking vision models under CLIP setting with an ImageNet-22K data scale.
    \item Sec.~\ref{sec:datacomp1b}: The numerical results of benchmarking vision models under CLIP setting with DataComp-1B.
    \item Sec.~\ref{sec:38dataset}: Detailed results of 38 datasets for different VLMs. 
    \item Sec.~\ref{sec:training_hyper}: The training hyper-parameter settings for short/long schedules and high-resolution input fine-tuning.
\end{itemize}

\section{Ablation Studies}
\label{sec:abl}

We conduct ablation studies on \modelname design from two aspects: macro-level network and micro-level block.
At the macro-level network design, we ablate the hybrid architecture and channel sizes of our three-stage network.
At the micro-level block design, we ablate the design choices of convolution blocks and feed-forward network. In the tables, `IN acc.' and `avg. 38' denote the ImageNet accuracy (\%) and the average accuracy (\%) of 38 datasets, respectively.
The ImageNet accuracy is used as the main metric.
For simplicity, all the ablation studies are performed using base model variants with 128M seen samples.

\textbf{Hybrid Architecture:}
In \tabref{tab:hybrid}, we ablate design choices of hybrid architectures.
Specifically, the compared architectures include ViT-B/16 (pure transformer with TFB or TFB-GeGLU blocks in stage 3), a new MBConvNet-B (pure ConvNet with MBConv-LN blocks in all three stages), and our \modelname-B (MBConv-LN in stage 1 and 2, and TFB-GeGLU in stage 3).
The ablated models may differ in depth but share a similar number of parameters.
As shown in the table, our \modelname-B outperforms both the pure Transformer ViT-B/16 and the pure ConvNet MBConvNet-B by more than +4.7\%.

\textbf{Channel Sizes of \modelname:}
We ablate the effect of varying channel sizes within our \modelname. The channel sizes ($x_1C$, $x_2C$, $x_3C$) denote the channel sizes of stage 1, 2, and 3, respectively.
We set the channel size multipliers $x_1$ and $x_2$ to be 1 and 2 (commonly used in the literature for ImageNet).
We ablate different values for $x_3$ in \tabref{tab:channel}. Our final setting of ($C$, $2C$, $6C$) improves over ($C$, $2C$, $8C$) by +1.1\%, and is on par with ($C$, $2C$, $4C$) but uses fewer parameters and MACs.

\begin{table}[t!]
    \centering
    \resizebox{\columnwidth}{!}{
    \begin{tabular}{c|c|c|c|c|c|c}
        & \multicolumn{2}{c|}{\# block type}  & depth & params & IN acc. & avg. 38\\
     model    &  stage 1 \& 2 & stage 3  &  stage 3  &  (M)  & (\%) &  datasets \\
        \hline
        ViT-B/16  & - & TFB &   12 & 86.2  & 45.8 & 41.0 \\
        ViT-B/16  & - & TFB-GeGLU &   14 & 84.2  & 45.4 & 40.9 \\
        ViT-B/16  & - & TFB-GeGLU &   15 & 90.2  & 46.1 & 41.2 \\
        MBConvNet-B & MBConv-LN & MBConv-LN &   18 & 87.3  & 45.8 & 41.7 \\
        \movittable{\modelname-B}  & \movittable{MBConv-LN} & \movittable{TFB-GeGLU} & \movittable{14} & \movittable{87.5}   & \movittable{50.8} & \movittable{44.6} \\
    \end{tabular}
    }
    \caption{
    \textbf{Ablation study for hybrid architecture.}
    MBConv-LN: Mobile Convolution with LayerNorm. TFB-GeGLU: Transformer Block with GeGLU.
    In this ablation study, we ablate TFB and TFB-GeGLU in ViT-B/16, and design a pure ConvNet using only MBConv-LN across all three stages (called MBConvNet-B in the table).
    Our final setting is marked in blue.
    }
    \label{tab:hybrid}
\end{table}

\begin{table}[t!]
    \centering
    \resizebox{0.86\columnwidth}{!}{
    \begin{tabular}{c|c|c|c|c}
    
        channel size  & params (M) & MACs (G) & IN acc. & avg. 38 \\
        \shline
        (C, 2C, 8C) & 86.0 & 19.5  & 49.7 & 44.8 \\
        \movittable{(C, 2C, 6C)} & \movittable{87.5} & \movittable{21.8}  & \movittable{50.8} & \movittable{44.6} \\
        (C, 2C, 4C) & 91.5 & 28.5  & 51.0 & 44.8 \\
    \end{tabular}
    }
    \caption{
    \textbf{Ablation study on the channel sizes.}
    The channel sizes ($x_1C$, $x_2C$, $x_3C$) denote the channel sizes of stage 1, 2, and 3, respectively, \wrt a constant C (\eg, $(x_1,x_2,x_3) = (1, 2, 6)$ and $C=128$ for \modelname-B).
    Our final setting is marked in blue.
    }
    \label{tab:channel}
\end{table}

\begin{table}[t!]
    \centering
    \resizebox{0.9\columnwidth}{!}{
    \begin{tabular}{c|c|c|c|c}
    
        block type  & params (M) & MACs (G) & IN acc. & avg. 38 \\
        \shline
        ConvNeXt &  88.0 & 21.0   & 49.8 & 44.9 \\
        MBConv-BN & 87.5 & 21.9   & 50.5 & 44.9 \\
        MBConv-BN-SE & 88.5 & 21.9   & 50.9 & 45.0 \\
        \movittable{MBConv-LN} &  \movittable{87.5} & \movittable{21.8}  & \movittable{50.8} & \movittable{44.6} \\
        
    \end{tabular}
    }
    \caption{
    \textbf{Ablation study for design choice of convolutional blocks.} BN: BatchNorm. SE: Squeeze-and-Exciation. LN: LayerNorm.
    Our final setting is marked in blue. 
    }
    \label{tab:abl:block}
\end{table}

\textbf{Design Choice of Convolution Blocks:}
In \tabref{tab:abl:block}, we ablate the design choices of convolution blocks in stage 1 and 2. The design choices include ConvNeXt, MBConv-BN, MBConv-BN-SE, and our MBConv-LN.
MBConv-BN block is the original MBConv block used in MobileNetv2~\cite{sandler2018mobilenetv2} with three BatchNorm layers~\cite{ioffe2015batch}, while the MBConv-BN-SE block, proposed by MobileNetv3~\cite{howard2019searching}, augments MBConv-BN with the Squeeze-and-Excitation layer~\cite{hu2018squeeze}.
Each of the MBConv variants demonstrates a superior performance to the ConvNeXt block~\cite{liu2022convnet}.
Our MBConv-LN, which employs a single Layer Normalization~\cite{ba2016layer}, outperforms the MBConv-BN block, and achieves a similar result to MBConv-BN-SE while requiring fewer parameters.

\textbf{Design Choice of Feed-Forward Network}:
In \tabref{tab:ffn}, we study the effectiveness of GeGLU~\cite{shazeer2020glu} in a Transformer Block (TFB)~\cite{vaswani2017attention}.
We experiment with ViT-B/16 and our \modelname-B, and ablate on the effect of using the original TFB \vs. the adopted TFB-GeGLU~\cite{shazeer2020glu}.
Remarkably, with the same depth of 12 blocks, \modelname-B with GeGLU can achieve 49.9\% accuracy and surpass the plain ViT-B/16 by a significant +4.1\% margin and requires 13\% fewer parameters.
Adding two more blocks to align the parameters with ViT-B/16, our \modelname-B boosts its performance to 50.8\%, which not only improves over the GeGLU-absent \modelname-B counterpart (last row) by +0.4\% but also maintains a reduced parameters by 26\%.

\begin{table}[t!]
    \centering
    \resizebox{\columnwidth}{!}{
    \begin{tabular}{c|c|c|c|c|c}

        image encoder  & GeGLU~\cite{shazeer2020glu} & depth & params (M)  & IN acc. & avg. 38 \\
        \shline
        ViT-B/16 &  & 12 & 86.2  & 45.8 & 41.0 \\
        \modelname-B & & 12 & 89.8  & 50.3 & 44.0 \\
        \modelname-B & \checkmark & 12 & 75.7  & 49.9 & 43.5 \\
        \movittable{\modelname-B} & \movittable{\checkmark} & \movittable{14} & \movittable{87.5}  & \movittable{50.8} & \movittable{44.6} \\
        \modelname-B &  & 14 & 104.0  & 50.4 & 44.5 \\
    \end{tabular}
    }
    \caption{
    \textbf{Ablation study for design choice of FFN.} 
    Our final setting is marked in blue.
    }
    \label{tab:ffn}
\end{table}

\section{Open-Vocabulary Dense Prediction}
\label{sec:ovlvis}
\textbf{Frozen Feature Extraction via Sliding Window:} 
We tested the transferability of VLMs to open-vocabulary detection tasks using F-ViT~\cite{wu2023clipself} and open-vocabulary segmentation tasks using FC-CLIP~\cite{yu2023convolutions} frameworks, which both rely on a frozen CLIP backbone. The image size (\eg, $1344\times1344$) for dense prediction tasks is usually larger than that of upstream VLM pre-training (\eg, $224\times224$). To employ a frozen transformer-based architecture in these framework, we did not use any distillation~\cite{wu2023clipself} or convolutional backbone~\cite{yu2023convolutions}, while we find that a simple sliding window strategy~\cite{yu2023towards} for frozen image feature extraction is effective enough to obtain reasonable performance on downstream tasks requiring high resolution input. The window size is the same as the input image size used during its VLM pretraining. We denote the slightly modified frameworks as \textit{Sliding F-ViT} and \textit{Sliding FC-CLIP}.  We follow~\cite{wu2023clipself, yu2023convolutions} and use $896\times896$ and $1344\times1344$ input size for the open-vocabulary detection and segmentation tasks, respectively. 

\subsection{Open-Vocabulary Detection}

In Tab.5 of main paper, ViTamin has been validated to be effective for open-vocabulary object detection on the OV-COCO dataset. In this section, we supplement the results on an additional benchmark OV-LVIS, where 
 ViTamin sets a new state-of-the-art performance.

\textbf{Experimental Setting:} The open-vocabulary LVIS (OV-LVIS), introduced in ViLD~\cite{gu2021open}, redefines the 337 rare categories from the LVIS v1.0~\cite{gupta2019lvis} dataset as novel categories. We strictly follow the F-ViT~\cite{wu2023clipself} framework to perform the open-vocabulary detection tasks, excepting the frozen image features are extracted in a sliding-window manner~\cite{yu2023towards} (denoted as \textit{Sliding F-ViT} in \tabref{tab:sota-lvis}). The effectiveness of VLMs is validated through simply replacing the frozen backbone of F-ViT~\cite{wu2023clipself} framework. For evaluation, we follow previous works to use the mean mask AP on rare categories (AP$_{r}$) as the metric on OV-LVIS.

\textbf{Results Analysis:}  \tabref{tab:open-voc-det:ovlvis} demonstrates that \modelname-L is a stronger image encoder for open-vocabulary detector, surpassing its ViT-L/14 counterpart by 3.1\% on OV-LVIS dataset \cite{gu2021open}. 


\begin{table}[!t]
\small
\centering
\scalebox{0.95}{
\tablestyle{4pt}{1.05}
\begin{tabular}{c|cc|c}
 \multirow{2}{*}{image encoder} &  \multicolumn{2}{c|}{pretraining}   & OV-LVIS~\cite{gu2021open}  \\
  & dataset & scheme  & (mAP$_{r}$) \\
\shline
ViT-L/14 & DataComp-1B & CLIPA-v2    & 32.5 \\
\textcolor{siglipcolor}{ConvNeXt-L} & \textcolor{siglipcolor}{LAION-2B} & \textcolor{siglipcolor}{OpenCLIP}  &  \textcolor{siglipcolor}{29.1}  \\
\hline 
\modelname-L & DataComp-1B & OpenCLIP   &  \textbf{35.6}  \\
\end{tabular}
}
\vspace{-1mm}
\caption{   
    \textbf{Open-vocabulary detection.}
    Different image encoders (ViT-L/14 by~\cite{li2023clipa} and ConvNeXt-L by~\cite{ilharco_gabriel_2021_5143773}) are deployed using the F-ViT framework~\cite{wu2023clipself} in a sliding window manner~\cite{yu2023towards}, trained on OV-LVIS dataset~\cite{gu2021open}. ConvNeXt-L is marked in gray due to different pretrained dataset.
}
\label{tab:open-voc-det:ovlvis}
\end{table}

\begin{table}[t!]
\centering
\scalebox{0.95}{
\tablestyle{4pt}{1.05}
\begin{tabular}{c|c|cc}
 \multirow{2}{*}{detector} & image  & OV-LVIS & OV-COCO\\
  & encoder & (AP$_{r}$) & (AP$_{50}^{\mathrm{novel}}$) \\
\shline
ViLD~\cite{gu2021open} & RN50 & 16.6 & 27.6 \\
OV-DETR~\cite{zang2022open} & RN50 & 17.4 & 29.4 \\
DetPro~\cite{du2022learning} & RN50 & 19.8 & - \\
OC-OVD~\cite{bangalath2022bridging} & RN50 & 21.1 & 36.6 \\
OADP~\cite{wang2023object} & RN50 & 21.7 & - \\
RegionCLIP~\cite{zhong2022regionclip} & RN50x4 & 22.0 & -\\
CORA~\cite{wu2023cora} & RN50x4 & 22.2 & 41.7 \\
BARON-KD~\cite{wu2023aligning} & RN50 & 22.6 & 34.0 \\
VLDet~\cite{lin2022learning} & SwinB & 26.3 & - \\
F-VLM~\cite{kuo2022f} & RN50x64 & 32.8 & 28.0 \\
Detic~\cite{zhou2022detecting} & SwinB & 33.8 & - \\
RO-ViT~\cite{kim2023region} & ViT-L/16 & 32.4 & 33.0 \\
RO-ViT~\cite{kim2023region} & ViT-H/16 & 34.1 & - \\
\hline
\hline
F-ViT~\cite{wu2023clipself} & ViT-L/14 & 24.2 & 24.7  \\
F-ViT+CLIPSelf~\cite{wu2023clipself} & ViT-L/14 & 34.9 & \textbf{44.3} \\
\hline
\ovdetmodelname & \textbf{\modelname-L} & \textbf{35.6} & 37.5 \\

\end{tabular}
}
\caption{\textbf{Comparison with prior arts} on open-vocabulary detection on OV-LVIS~\cite{gu2021open} and OV-COCO~\cite{zareian2021open}. 
The last row (\ovdetmodelname) shows the result of employing our \modelname-L using the F-ViT framework~\cite{wu2023clipself} in a sliding window manner~\cite{yu2023towards}.
}
\label{tab:sota-lvis}
\end{table}

\textbf{Comparison with Prior Arts:} 
As shown in~\tabref{tab:sota-lvis}, \modelname consistently outperforms all previous methods in the open-vocabulary detection task on OV-LVIS, setting a new state-of-the-art performance of 35.6\% AP$_{r}$. Notably, our approach surpasses not only the distillation-based backbone (\eg, CLIPSelf~\cite{wu2023clipself}) but also larger backbone (\eg, ViT-H/16 in RO-ViT~\cite{kim2023region}).

\subsection{Open-Vocabulary Segmentation}

In Tab.6 of main paper, \modelname has been validated to be effective for open-vocabulary panoptic and semantic segmentation on 8 dataset. We strictly follow the FC-CLIP framework~\cite{yu2023convolutions}  to perform the open-vocabulary segmentation tasks, excepting the frozen image features are extracted in a sliding-window manner~\cite{yu2023towards} (denoted as \textit{Sliding FC-CLIP} in \tabref{tab:open-voc-det:ovlvis}).
Following prior works~\cite{yu2023convolutions}, the \textit{Sliding FC-CLIP} is trained on COCO~\cite{lin2014microsoft} and zero-shot evaluated on the other datasets. In this section, we compare \modelname with previous state-of-the-art methods. 

\textbf{Comparison with Prior Arts:}
In \tabref{tab:open-voc-seg:sota}, our approach consistently outperforms all previous open-vocabulary segmentation methods in 2 panoptic dataset and 4 semantic benchmarks, setting a new state-of-the-art. Notably, \modelname surpasses the the prior art by 0.5\% PQ on ADE panoptic dataset and 1.5\% mIOU on A-150 semantic dataset.

\begin{table}[!t]
\small
\centering
\scalebox{0.74}{
\tablestyle{2pt}{1.1}
\begin{tabular}{c|c|ccc|ccccc}
                           & & \multicolumn{3}{c|}{panoptic dataset (PQ)}   & \multicolumn{5}{c}{semantic dataset (mIoU)}                                \\
method & image   & ADE            & Cityscapes   & MV   & A-150 &A-847 & PC-459 & PC-59 & PAS-21      \\
 & encoder  & \cite{zhou2017scene} & \cite{cordts2016cityscapes} & \cite{neuhold2017mapillary} &  \cite{zhou2017scene} & \cite{zhou2017scene} & \cite{mottaghi2014role} & \cite{mottaghi2014role} & \cite{everingham2010pascal}  \\
\shline
FreeSeg~\cite{qin2023freeseg} & -   & 16.3 & -  & - & - & - & - & - & -  \\
OpenSeg~\cite{ghiasi2022scaling} & -   & - & -  & - & 21.1 & 6.3 & 9.0 & 42.1 & -  \\
GroupViT~\cite{xu2022groupvit} & ViT-S/16   & - & -  & - & 10.6 & 6.3 & 9.0 & 42.1 & -  \\
MaskCLIP~\cite{dong2023maskclip} & ViT-B/16   & 15.1 & -  & - & 23.7 & 8.2 & 10.0 & 45.9 & -  \\
ODISE~\cite{xu2023open} &  - & 22.2 & 23.9 & 14.2 & 29.9 & 11.1 & 14.5 & 57.3 & \textbf{84.6}  \\
\hline\hline
FC-CLIP~\cite{yu2023convolutions} & ConvNeXt-L   & 26.8 & 44.0 & \textbf{18.3} & 34.1 & 14.8 & 18.2 & 58.4 & 81.8 \\
\hline 
\ovsegmodelname & \textbf{\modelname-L}  & \textbf{27.3} & \textbf{44.0} & 18.2 & \textbf{35.6} & \textbf{16.1} & \textbf{20.4} & \textbf{58.4} & 83.4  \\
\end{tabular}
}
\caption{
    \textbf{Comparison with prior arts} on open-vocabulary segmentation. \modelname sets a new state-of-the-art result on various panoptic and semantic segmenation datasets. 
    The last row (\ovsegmodelname) shows the result of employing our \modelname-L using the FC-CLIP framework~\cite{yu2023convolutions} in a sliding window manner~\cite{yu2023towards}. 
}
\label{tab:open-voc-seg:sota}
\vspace{-2.5mm}
\end{table}

\section{Locked-Text Tuning}
\label{sec:appendix_ltt}
\tabref{tab:ltt} summarizes the detailed results of using the proposed new training scheme, Locked-Text Tuning (LTT). Specifically, when using the LTT training scheme, we employ the text encoder pretrained from \modelname-L, and use it to guide the training of image encoders of \modelname-S and \modelname-B.  As shown in the table, we consistently observe the improvements of using LTT. Compared to other distillation-based CLIP training schemes (See the rows marked in grey), our models achieve higher classification and retrieval accuracy in similar model parameters. Practically, despite being adopted from the larger model, the text encoder is much lighter compared to the image encoder (6.6 vs 21.8 GMACs), resulting in only a 14\% increase in overall model MACs.
Interestingly, using LTT results in a 10\% savings in training costs for \modelname-B, due to the text encoder being fully frozen.

\begin{table}[t!]
    \centering
    \resizebox{\columnwidth}{!}{
    \begin{tabular}{c|c|c|c|c|c|c|c}

        training  & training & image  & params  & seen  & IN acc. & avg. 38 & retrieval \\
      scheme  & dataset &  encoder &  (M) & samp. & (\%) & (\%) &  COCO (\%) \\
        \shline
        \multicolumn{8}{l}{\baseline{\textit{\textbf{models on private/other dataset, for reference}}}} \\ 
        \textcolor{siglipcolor}{LiT}~\cite{zhai2022lit}  & \textcolor{siglipcolor}{Private-4B} & \textcolor{siglipcolor}{ViT-B/32} & \textcolor{siglipcolor}{86.2} & \textcolor{siglipcolor}{0.9B} &  \textcolor{siglipcolor}{68.8} & \textcolor{siglipcolor}{-} & \textcolor{siglipcolor}{36.1} \\ 
        \textcolor{siglipcolor}{TinyCLIP}~\cite{wu2023tinyclip} & \textcolor{siglipcolor}{LAION+YFCC} & \textcolor{siglipcolor}{ViT-45M/32} & \textcolor{siglipcolor}{45.0} & \textcolor{siglipcolor}{1.6B} & \textcolor{siglipcolor}{62.1} & \textcolor{siglipcolor}{-} &  \textcolor{siglipcolor}{45.4} \\
        \textcolor{siglipcolor}{TinyCLIP}~\cite{wu2023tinyclip} & \textcolor{siglipcolor}{LAION+YFCC} & \textcolor{siglipcolor}{ViT-63M/32} & \textcolor{siglipcolor}{63.0} & \textcolor{siglipcolor}{1.6B} & \textcolor{siglipcolor}{64.5} & \textcolor{siglipcolor}{-} & \textcolor{siglipcolor}{47.7} \\
        \shline
        \multicolumn{8}{l}{\baseline{\textit{\textbf{our experiments}}}} \\ 
        OpenCLIP & DataComp-1B & \modelname-S & 22.0 & 128M & 43.3 & 40.8 & 25.8 \\
        OpenCLIP & DataComp-1B & \modelname-S & 22.0 & 512M & 57.3 & 49.6  & 36.6 \\
        OpenCLIP & DataComp-1B & \modelname-S & 22.0 & 1.28B & 62.2 & 53.2 & 40.2 \\
        \hline
        OpenCLIP & DataComp-1B & \modelname-B & 87.5 & 128M & 50.8 & 44.6 & 31.2 \\
        OpenCLIP & DataComp-1B & \modelname-B & 87.5 & 512M & 64.0 & 53.9 & 41.7 \\
        OpenCLIP & DataComp-1B & \modelname-B & 87.5 & 1.28B & 68.9 & 57.7 & 44.9 \\
        \hline \hline
        LTT (ours) & DataComp-1B & \modelname-S & 22.0 & 128M & 47.5 & 44.8 & 33.4 \\
        LTT (ours) & DataComp-1B & \modelname-S & 22.0 & 512M & 58.9 & 52.0 & 41.6 \\
        LTT (ours) & DataComp-1B & \modelname-S & 22.0 & 1.28B & 63.4 & 54.6 & 45.0 \\
        \hline
        LTT (ours) & DataComp-1B & \modelname-B & 87.5 & 128M & 56.7 & 50.5 & 39.8 \\
        LTT (ours) & DataComp-1B & \modelname-B & 87.5 & 512M & 66.8 & 57.3 & 47.1 \\
        LTT (ours) & DataComp-1B & \modelname-B & 87.5 & 1.28B & 70.8 & 59.4 & 50.0 \\
        
    \end{tabular}
    }
    \caption{
    \textbf{Locked-Text Tuning (LTT) training scheme.}
    We use the pretrained text encoder from \modelname-L and train the image encoders of \modelname-\{S,B\}.
    Due to the use of private or other filtered/merged dataset, the results borrowed from LiT~\cite{zhai2022lit} and TinyCLIP~\cite{wu2023tinyclip} are just for reference, and  LiT~\cite{zhai2022lit} reports retrieval on COCO only. $\dagger$: a filtered subset of WebLI dataset~\cite{chen2022pali}. 
    }
    \label{tab:ltt}
\end{table}

\section{Benchmarking Vision Models in CLIP with ImageNet-22K Data Scale}
\label{sec:imagenet22k}
\tabref{tab:imagenet_scale} summarizes the results of benchmarking vision models under CLIP setting with ImageNet-22K data scale.
Specifically, we mimic the ImageNet-22K data scale by randomly selecting 14.2M data samples from DataComp-1B, and set the training epochs to 90, a standard training setting on ImageNet-22K.
Similar to the findings on ImageNet-22K in the literature~\cite{liu2022convnet}, under such a small data scale (14.2M data samples), ConvNeXt-T consistently outperforms ViT-S/32 and ViT-S/16.
However, when the data scales up to 128M, or even 1.28B, the results are totally different, where ViT/16 shows a superior performance to ConvNeXt by a large margin, across all model sizes (see~\tabref{tab:benchmark}).
We note that hybrid models, such as CoAtNet-0 and our \modelname-S, still demonstrate the best performances under this small data scale, showing that the hybrid design works well across all data sizes.

\begin{table}[t!]
\centering
\scalebox{0.7}{
\begin{tabular}{c|c|c|c|c|c|c}
 image & data & epoch & \#params & MACs  & ImageNet & avg. 38  \\
   encoder & size (M) &   &  (M) &  (G) &Acc.(\%) & datasets (\%)   \\
\shline
\multicolumn{7}{l}{\baseline{\textit{\textbf{ImageNet-22K scale}}}} \\
 ViT-S/32   & 14.2  & 90  & 21.81 & 1.12 & 39.4 & 36.7   \\
 ViT-S/16    & 14.2  & 90  & 21.81  & 4.25 & 45.7 & 38.7  \\
ConvNeXt-T   & 14.2  & 90   & 28.61 & 4.47 & 45.9 & 39.3   \\
 CoAtNet-0   & 14.2  & 90  & 24.56 & 4.43  & 49.1 & 41.4   \\
\hline
\modelname-S  & 14.2  & 90   & 22.03  & 5.50 & 50.3 & 41.3 
  
\end{tabular}
}
\caption{
\textbf{Benchmarking vision models under CLIP setting with an ImageNet-22K data scale.}
We mimic the ImageNet-22K data scale with 14.2M data size and 90 training epochs (standard training setting on ImageNet-22K). The benchmarked vision models include ViT (pure transformer), ConvNeXt (pure convolution), CoAtNet (hybrid model), and our proposed \modelname.
}
\label{tab:imagenet_scale}
\end{table}

\begin{table*}[ht]
\setlength{\tabcolsep}{1.5mm} 
\centering
\resizebox{1\textwidth}{!}{%
\begin{tabular}{c|c|c|c|c|c|c|c|c|c|c}
 & image  & text encoder & seen   & \#params (M) & MACs (G)   & ImageNet & avg. 38 & ImageNet & VTAB & Retrieval  \\

image encoder & size & depth / width  & samples   & image+text & image+text & Acc. (\%) & datasets & dist. shift. &  &   \\
\shline
\shline
\multicolumn{11}{l}{\baseline{\textit{\textbf{small model variants}}}} \\
 ViT-S/32 & 224 & 12 / 384 & 128M   & 21.81 + 40.44 & 1.12 + 1.64  & 32.1 & 34.1  & 25.3 & 35.8 & 27.2  \\
 ViT-S/16 & 224 & 12 / 384 & 128M   & 21.81 + 40.44 & 4.25 + 1.64  & 38.4 & 36.7 & 29.3 & 38.5 & 31.3  \\
ConvNeXt-T & 224 & 12 / 384 & 128M  & 28.61 + 40.44 & 4.47 + 1.64  & 37.0 & 35.8 & 30.1 & 37.5 & 30.8  \\
 CoAtNet-0 & 224 & 12 / 384 & 128M  & 24.56 + 40.44 & 4.43 + 1.64  & 42.4 & 38.9 & 33.5 & 39.9 & 34.2  \\
 \hline
  \modelname-S  & 224 & 12 / 384 & 128M & 22.03 + 40.44 & 5.50 + 1.64  & \textbf{43.3} & \textbf{40.8} & \textbf{35.6} & \textbf{41.0} & \textbf{35.2}  \\
\hline \hline
 ViT-S/32 & 224 & 12 / 384 & 512M & 21.81 + 40.44 & 1.12 + 1.64 & 47.3 & 44.3 & 36.7 & 46.9 & 36.9  \\
 ViT-S/16 & 224 & 12 / 384 & 512M   & 21.81 + 40.44 & 4.25 + 1.64  & 53.8 & 46.5 & 41.9 & 46.7 & 42.1  \\
ConvNeXt-T & 224 & 12 / 384 & 512M  & 28.61 + 40.44 & 4.47 + 1.64  & 53.3 & 46.1 & 42.4 & 46.4 & 42.2  \\
 CoAtNet-0 & 224 & 12 / 384 & 512M  & 24.56 + 40.44 & 4.43 + 1.64  & 56.4 & 49.0 & 45.2 & \textbf{49.0} & 45.0  \\
 \hline
  \modelname-S  & 224 & 12 / 384 & 512M & 22.03 + 40.44 & 5.50 + 1.64  & \textbf{57.3} & \textbf{49.6} & \textbf{46.9} & 48.8 & \textbf{45.4}  \\
 \hline \hline
 ViT-S/32 & 224 & 12 / 384 & 1.28B & 21.81 + 40.44 & 1.12 + 1.64  & 53.0 & 47.3 & 41.3 & 48.9  & 41.5  \\
 ViT-S/16 & 224 & 12 / 384 & 1.28B   & 21.81 + 40.44 & 4.25 + 1.64  & 59.8 & 50.9 & 47.2 & 51.3 & 47.5 \\
ConvNeXt-T & 224 & 12 / 384 & 1.28B  & 28.61 + 40.44 & 4.47 + 1.64  & 59.9 & 51.3 & 47.8 & 52.7 & 48.3  \\
 CoAtNet-0 & 224 & 12 / 384 & 1.28B  & 24.56 + 40.44 & 4.43 + 1.64  & 61.7 & 51.6 & 50.1 & 51.3 & 48.9  \\
 \hline
  \modelname-S  & 224 & 12 / 384 & 1.28B & 22.03 + 40.44 & 5.50 + 1.64  & \textbf{62.2} & \textbf{53.2} & \textbf{51.3} & \textbf{51.7} & \textbf{50.0}  \\ 
\shline
\multicolumn{11}{l}{\baseline{\textit{\textbf{base model variants}}}} \\
   ViT-B/32 & 224 & 12 / 512 & 128M  & 86.19 + 63.43 & 4.37 + 2.91  & 38.9 & 38.0 & 30.6 & 40.6 & 30.7  \\
   ViT-B/16 & 224 & 12 / 512 & 128M  & 86.19 + 63.43 & 16.87 + 2.91  & 45.8 & 41.0 & 35.8 & 42.1 & 36.2  \\

 ConvNeXt-B & 224 & 12 / 512 & 128M  & 88.09 + 63.43 & 15.38 + 2.91  & 41.4 & 39.7 & 33.5 & 41.2 & 34.1  \\

CoAtNet-2 & 224 & 12 / 512 & 128M  & 74.18 + 63.43 & 15.94 + 2.91  & 48.5 & 43.5 & 38.9 & 43.8 & 39.1  \\
\hline
 \modelname-B  & 224 & 12 / 512 & 128M & 87.53 + 63.43 & 21.84 + 2.91  &  \textbf{50.8} & \textbf{44.6} & \textbf{41.3} & \textbf{45.1} & \textbf{40.8}  \\

 \hline \hline
 ViT-B/32 & 224 & 12 / 512 & 512M  & 86.19 + 63.43 & 4.37 + 2.91  & 54.8 & 48.3 & 42.7 & 50.1 & 42.4  \\
 ViT-B/16 & 224 & 12 / 512 & 512M  & 86.19 + 63.43 & 16.87 + 2.91  & 60.0 & 51.0 & 48.2 & 51.4 & 47.5  \\

ConvNeXt-B & 224 & 12 / 512 & 512M  & 88.09 + 63.43 & 15.38 + 2.91  & 59.4 & 50.3 & 47.9 & 49.9 & 47.2  \\

CoAtNet-2 & 224 & 12 / 512 & 512M  & 74.18 + 63.43 & 15.94 + 2.91  & 63.3 & 52.4 & 52.4 & 51.0 & 49.7  \\
\hline
\modelname-B  & 224 & 12 / 512 & 512M & 87.53 + 63.43 & 21.84 + 2.91  & \textbf{64.0} & \textbf{53.9} & \textbf{53.3} & \textbf{53.7} & \textbf{50.8}  \\
\hline \hline
 ViT-B/32 & 224 & 12 / 512 & 1.28B  & 86.19 + 63.43 & 4.37 + 2.91  & 60.1 & 52.5 & 47.4 & 53.6 & 47.5  \\
 ViT-B/16 & 224 & 12 / 512 & 1.28B  & 86.19 + 63.43 & 16.87 + 2.91  & 65.6 & 55.6 & 53.1 & 55.3 & 51.7  \\
 ConvNeXt-B & 224 & 12 / 512 & 1.28B  & 88.09 + 63.43 & 15.38 + 2.91  & 65.3 & 54.7 & 54.0 & 54.2 & 51.7  \\

CoAtNet-2 & 224 & 12 / 512 & 1.28B  & 74.18 + 63.43 & 15.94 + 2.91  & 68.5 & 56.8 & 57.2 & 56.0 & 53.4  \\

\hline
\modelname-B  & 224 & 12 / 512 & 1.28B & 87.53 + 63.43 & 21.84 + 2.91  & \textbf{68.9} & \textbf{57.7} & \textbf{58.3} & \textbf{56.4} & \textbf{54.1}  \\

\shline
\multicolumn{11}{l}{\baseline{\textit{\textbf{large model variants}}}} \\
 ViT-L/32 & 224 & 12 / 768 & 128M  & 303.97 + 123.65 & 15.27 + 6.55  & 43.5 & 40.8 & 34.0 & 42.7 & 34.2  \\
 ViT-L/16 & 224 & 12 / 768 & 128M  & 303.97 + 123.65 & 59.70 + 6.55  & 49.4 & 43.8 & 38.7 & 44.3 & 38.9  \\
 ViT-L/14 & 224 & 12 / 768 & 128M  & 303.97 + 123.65 & 77.83 + 6.55  & 49.9 & 43.8 & 39.4 & 44.5 & 39.3  \\
ConvNeXt-XL & 224 & 12 / 768 & 128M  & 350.25 + 123.65 & 79.65 + 6.55  & 42.8 & 38.4 & 33.3 & 38.4 & 35.0  \\
CoAtNet-4 & 224 & 12 / 768 & 128M  & 275.07 + 123.65 & 60.81 + 6.55  & 52.5 & \textbf{45.2} & 42.0 & \textbf{45.2} & 41.1  \\
\hline
\modelname-L  & 224 & 12 / 768 & 128M & 333.32 + 123.65 & 72.60 + 6.55  & \textbf{52.7} & 44.8 & \textbf{42.4} & 44.6 & \textbf{41.8}  \\

\hline \hline
 ViT-L/32 & 224 & 12 / 768 & 512M  & 303.97 + 123.65 & 15.27 + 6.55  & 60.4 & 51.8 & 47.4 & 52.7 & 47.3  \\
ViT-L/16 & 224 & 12 / 768 & 512M  & 303.97 + 123.65 & 59.70 + 6.55  & 66.4 & 55.6 & 53.6 & 55.5 & 52.2  \\
 ViT-L/14 & 224 & 12 / 768 & 512M  & 303.97 + 123.65 & 77.83 + 6.55  & 67.0 & 55.4 & 54.8 & 54.2 & 52.0  \\
ConvNeXt-XL & 224 & 12 / 768 & 512M  & 350.25 + 123.65 & 79.65 + 6.55  & 63.0 & 52.5 & 51.1 & 51.8 & 49.4  \\
CoAtNet-4 & 224 & 12 / 768 & 512M  & 275.07 + 123.65 & 60.81 + 6.55  & 66.8 & 56.1 & 56.4 & \textbf{56.5} & 50.4  \\

\hline
\modelname-L  & 224 & 12 / 768 & 512M & 333.32 + 123.65 & 72.60 + 6.55  & \textbf{68.7} & \textbf{56.6} & \textbf{56.8} & \textbf{56.5} & \textbf{53.2}  \\

 \hline \hline
 ViT-L/32 & 224 & 12 / 768 & 1.28B  & 303.97 + 123.65 & 15.27 + 6.55  & 67.5 & 57.0 & 54.1 & 57.9 & 51.9  \\
 ViT-L/16 & 224 & 12 / 768 & 1.28B  & 303.97 + 123.65 & 59.70 + 6.55  & 71.9 & 60.1 & 59.9 & 59.9 & 56.0  \\
 ViT-L/14 & 224 & 12 / 768 & 1.28B  & 303.97 + 123.65 & 77.83 + 6.55  & 72.3 & 60.7 & 60.5 & 60.0 & 56.0  \\
ConvNeXt-XL & 224 & 12 / 768 & 1.28B  & 350.25 + 123.65 & 79.65 + 6.55  & 70.2 & 58.3 & 59.1 & 57.0 & 55.5  \\
CoAtNet-4 & 224 & 12 / 768 & 1.28B  & 275.07 + 123.65 & 60.81 + 6.55  & 71.3 & 59.4 & 61.4 & 59.1 & 53.4  \\
\hline
\modelname-L  & 224 & 12 / 768 & 1.28B & 333.32 + 123.65 & 72.60 + 6.55  & \textbf{73.9} & \textbf{62.0} & \textbf{62.9} & \textbf{61.4} & \textbf{56.6}  
\end{tabular}
}
\caption{
\textbf{Benchmarking vision backbones on Datacomp-1B under CLIP setting (contrastive language-image pretraining).}
We benchmark popular vision backbones, including ViT~\cite{dosovitskiy2020image} (pure transformer model), ConvNeXt~\cite{liu2022convnet} (pure convolution model), CoAtNet~\cite{dai2021coatnet} (hybrid convolution-transformer model), and our proposed \modelname, under different model parameters and training seen samples.
}
\label{tab:benchmark}
\end{table*}

\begin{table*}[!t]

\setlength{\tabcolsep}{0.4mm} 
\centering
\resizebox{1.02\textwidth}{!}{
\begin{tabular}{cc|c|ccccccccccccccccccccccccccccccccccccccccc}
\rotatebox{90}{image encoder}  & \rotatebox{90}{training  scheme}  & \rotatebox{90}{avg. 38} & \rotatebox{90}{ImageNet 1k~\cite{deng2009imagenet}} & \rotatebox{90}{Caltech-101~\cite{fei2004learning}} & \rotatebox{90}{CIFAR-10~\cite{krizhevsky2009learning}} & \rotatebox{90}{CIFAR-100~\cite{krizhevsky2009learning}} & \rotatebox{90}{CLEVR Counts~\cite{johnson2017clevr}} & \rotatebox{90}{CLEVR Distance~\cite{krizhevsky2009learning}} & \rotatebox{90}{Country211~\cite{radford2021learning}} & \rotatebox{90}{Describable Textures~\cite{cimpoi2014describing}} & \rotatebox{90}{EuroSAT~\cite{helber2019eurosat}} & \rotatebox{90}{FGVC Aircraft~\cite{maji2013fine}} & \rotatebox{90}{Food-101~\cite{bossard2014food}} & \rotatebox{90}{GTSRB~\cite{stallkamp2011german}} & \rotatebox{90}{ImageNet Sketch~\cite{wang2019learning}} & \rotatebox{90}{ImageNet v2~\cite{recht2019imagenet}} & \rotatebox{90}{ImageNet-A~\cite{hendrycks2021natural}} & \rotatebox{90}{ImageNet-O~\cite{hendrycks2021natural}} & \rotatebox{90}{ImageNet-R~\cite{hendrycks2021many}} & \rotatebox{90}{KITTI Vehicle Distance~\cite{geiger2012we}} & \rotatebox{90}{MNIST~\cite{lecun1998mnist}} & \rotatebox{90}{ObjectNet~\cite{barbu2019objectnet}} & \rotatebox{90}{Oxford Flowers-102~\cite{nilsback2008automated}} & \rotatebox{90}{Oxford-IIIT Pet~\cite{parkhi2012cats}} & \rotatebox{90}{Pascal VOC 2007~\cite{everingham2010pascal}} & \rotatebox{90}{PatchCamelyon~\cite{veeling2018rotation}} & \rotatebox{90}{Rendered SST2~\cite{zhai2019visual}} & \rotatebox{90}{RESISC45~\cite{zhai2019visual}} & \rotatebox{90}{Stanford Cars~\cite{krause20133d}} & \rotatebox{90}{STL-10~\cite{coates2011analysis}} & \rotatebox{90}{SUN397~\cite{xiao2016sun}} & \rotatebox{90}{SVHN~\cite{netzer2011reading}} & \rotatebox{90}{Flickr~\cite{young2014image}} & \rotatebox{90}{MSCOCO~\cite{chen2015microsoft}} & \rotatebox{90}{WinoGAViL~\cite{bitton2022winogavil}} & \rotatebox{90}{iWildCam~\cite{beery2021iwildcam}} & \rotatebox{90}{Camelyon17~\cite{bandi2018detection}} & \rotatebox{90}{FMoW~\cite{christie2018functional}} & \rotatebox{90}{Dollar Street~\cite{rojas2022dollar}} & \rotatebox{90}{GeoDE~\cite{ramaswamy2023beyond}}            \\

\shline
ViT-L/14 & \cite{ilharco_gabriel_2021_5143773} & 66.3 & 79.2 & 94.7 & 98.2 & 87.3 & 35.6 & 24.4 & 31.6 & 66.5 & 71.2 & 47.5 & 94.5 & 58.5 & 68.0 & 72.1 & 69.6 & 32.6 & 90.8 & 27.9 & 86.6 & 74.3 & 82.6 & 95.1 & 82.5 & 51.2  & 61.0 & 69.4 & 93.1 & 99.3 & 74.3 & 67.7 & 81.2 & 54.5 & 46.7 & 16.1 & 50.9 & 24.0 & 66.2 & 91.5 \\

ViT-L/14  & \cite{li2023clipa}  &  65.4 & 79.6 & 94.5 & 98.7 & 88.5 & 18.6 & 24.5 & 29.4 & 69.6 & 60.4 & 43.0 & 94.2 & 59.1 & 70.6 & 73.0 & 71.2 & 33.7 & 92.9 & 19.3 & 73.7 & 69.9 & 81.0 & 95.0 & 80.7 & 59.2 & 53.9 & 68.4 & 93.7 & 99.2 & 75.3 & 63.9 & 81.9 & 56.0 & 43.9 & 17.2 & 67.6 & 24.6 & 66.5 & 91.5  \\

ViT-L/14 $^\dagger$ & \cite{li2023clipa}  &
65.7 & 80.3 & 94.4 & 98.6 & 88.3 & 15.7 & 24.4 & 30.7 & 68.6 & 58.1 & 42.8 & 94.6 & 57.0 & 70.9 & 73.5 & 77.7 & 32.9 & 93.3 & 20.0 & 76.7 & 73.2 & 81.0 & 95.0 & 79.8 & 60.3 & 53.2 & 68.8 & 94.1 & 99.3 & 75.6 & 62.9 & 82.5 & 56.4 & 44.5 & 19.4 & 67.8 & 25.0 & 67.5 & 92.4 \\
\hline
\modelname-L  & \cite{ilharco_gabriel_2021_5143773} & 66.7 & 80.8 & 95.1 & 98.5 & 88.0 & 35.4 & 24.1 & 33.0 & 68.1 & 66.6 & 49.2 & 94.9 & 61.9 & 71.5 & 73.6 & 72.4 & 32.0 & 93.0 & 23.6 & 81.8 & 76.1 & 85.0 & 95.6 & 80.2 & 52.2 & 61.1 & 73.0 & 94.8 & 99.3 & 75.8 & 66.1 & 81.1 & 54.7 & 45.1 & 17.2 & 58.0 & 16.4 & 68.1 & 91.5 \\
\modelname-L$^\dagger$ & \cite{ilharco_gabriel_2021_5143773} & 67.2 & 81.8 & 95.6 & 98.5 & 87.8 & 31.7 & 24.1 & 36.0 & 69.2 & 64.7 & 49.7 & 95.8 & 63.4 & 72.1 & 75.2 & 81.7 & 31.1 & 93.8 & 21.2 & 81.3 & 80.7 & 84.7 & 95.8 & 82.0 & 50.1 & 60.5 & 73.1 & 95.1 & 99.5 & 76.3 & 66.9 & 82.5 & 55.7 & 47.3 & 19.1 & 49.5 & 17.5 & 70.7 & 92.2 \\
\modelname-XL$^\dagger$ & \cite{ilharco_gabriel_2021_5143773} & \textbf{68.1} & 82.6 & 95.7 & 98.7 & 88.8 & 19.1 & 20.0 & 37.8 & 71.5 & 75.6 & 53.7 & 96.0 & 53.2 & 73.1 & 76.3 & 83.1 & 33.0 & 94.2 & 17.4 & 88.9 & 81.9 & 85.6 & 95.9 & 83.8 & 56.2 & 61.9 & 75.9 & 94.8 & 99.4 & 76.2 & 74.0 & 84.7 & 58.7 & 47.9 & 21.4 & 46.0 & 22.5 & 68.6 & 92.5
\end{tabular}
}
\caption{
    \textbf{Detailed results of 38 dataset for different VLMs.} The compared models are trained with the scheme of either  OpenCLIP~\cite{ilharco_gabriel_2021_5143773} or CLIPA-v2~\cite{li2023clipa}. All models are trained on DataComp-1B~\cite{gadre2023datacomp} dataset with similar seen samples for a fair comparison. $\dagger$: using larger number of patches of 576 (\ie, image size of 336 for row 3 and 384 for row 5, respectively). 
}
\label{tab:38dataset}
\vspace{-4mm}
\end{table*}

\section{Numerical Results of Benchmarking Vision Models with DataComp-1B}
\label{sec:datacomp1b}

In Fig.2 of the main paper, we provide the analysis of benchmarked results from various aspects. In this section,  we further supplement the numerical results of benchmarking vision models (including ViT, ConvNeXt, CoAtNet, and our \modelname) across different model scales and data sizes in \tabref{tab:benchmark}.
As shown in the table, the proposed \modelname consistently outperforms all the other vision models in almost all settings.

\section{Results of 38 dataset for different VLMs.}
\label{sec:38dataset}

\tabref{tab:38dataset} demonstrates the detailed results for VLMs with different large-variant image encoders. This table is associated with Tab. 3 of the main paper.

\begin{table}[t!]
\tablestyle{5.0pt}{1.02}
\footnotesize
\begin{tabular}{@{\hskip -0.05ex}l|c@{\hskip 1ex}c}
& \textit{short schedule} & \textit{long schedule} \\
\multirow{2}{*}{training config} & \modelname-S/B/L &  \modelname-L/L2/XL/XL \\
& 224$^2$ & 224$^2$/224$^2$/256$^2$/256$^2$ \\
\shline
batch size & 8k/8k/16k & 90k \\
seen samples & 1.28B & 12.8B/12.8B/12.8B/40B \\
optimizer & AdamW & AdamW\\
base learning rate & 5e-4 & 2e-3 \\
weight decay & 0.02 & 0.02 \\
optimizer momentum $\beta_1$ & 0.9 & 0.9 \\
optimizer momentum $\beta_2$ & 0.98/0.98/0.95 & 0.95 \\
learning rate schedule & cosine decay & cosine decay \\
warmup steps & 500 & 782/4436/4436/9981 \\
warmup schedule & linear & linear \\
random crop ratio & none & [0.4, 1.0] \\
stochastic depth ~\cite{huang2016deep} & 0.1 & 0.1 \\
precision & amp bfloat16 & amp bfloat16
\end{tabular}
\caption{\textbf{Short/Long schedule training settings for \modelname variants}.}
\label{tab:train_detail}
\end{table}

\begin{table}[t!]
\tablestyle{2.0pt}{1.0}
\footnotesize
\begin{tabular}{@{\hskip -0.05ex}l|c@{\hskip 1ex}cc} 
\multirow{2}{*}{pre-training config} & \modelname-L & \modelname-L2 & \modelname-XL \\
& 224$^2$ & 224$^2$ & 256$^2$\\
\hline
fine-tuning config & 256$^2$/336$^2$/384$^2$ & 256$^2$/336$^2$/384$^2$ & 256$^2$/384$^2$ \\
\shline
batch size & 90k & 90k & 90k \\
seen samples & 0.2B & 0.5B & 0.5B \\
optimizer & AdamW & AdamW & AdamW \\
base learning rate & 1e-5 & 1e-5 & 1e-5 \\
weight decay & 0 & 0 & 0 \\
optimizer momentum $\beta_1$ & 0.9 & 0.9 & 0.9 \\
optimizer momentum $\beta_2$ & 0.95 & 0.95 & 0.95 \\
learning rate schedule & constant & constant & constant \\
warmup steps & 0 & 0 & 0  \\
random crop ratio & none & none & none  \\
stochastic depth ~\cite{huang2016deep} & 0.1 & 0.1 & 0.1 \\
precision & amp bfloat16 & amp bfloat16 & amp bfloat16 
\end{tabular}
\caption{\textbf{Fine-tuning setting for high resolution}. The models are pre-trained with \textit{long schedule} and then fine-tuned on the target resolution.
}
\label{tab:ft_detail}
\end{table}

\section{Training Hyper-parameter Settings}
\label{sec:training_hyper}

\tabref{tab:train_detail} and~\tabref{tab:ft_detail} provide our details of training hyper-parameter settings for short/long schedules and fine-tuning for high resolution, respectively.
The short schedule is used to benchmark several vision models on DataComp-1B, along with our ablation studies, while the long schedule is used to train our \modelname-L for better performances.
When fine-tuning the trained model on larger input resolution, we fine-tune with only 200M seen samples and a small constant learning rate.